\documentclass[10pt,onecolumn]{IEEEtran}

\usepackage[switch]{lineno}
\usepackage[cmex10]{amsmath}
\usepackage{amssymb,latexsym,amsfonts,stmaryrd}
\usepackage{amsthm}
\usepackage{mathtools}
\usepackage{graphicx}
\usepackage[noadjust]{cite}
\usepackage{algorithm}
\usepackage{algorithmic}
\usepackage{subfigure}
\usepackage{color}
\usepackage{soul}
\usepackage{lineno}
\usepackage{mathptmx}

\usepackage{etoolbox}
\makeatletter
\patchcmd{\@makecaption}
  {\scshape}
  {}
  {}
  {}
\makeatother

\def\tablename{Table}

\usepackage[hyphens]{url}
\usepackage[hidelinks]{hyperref}
\hypersetup{breaklinks=true}
\usepackage{cite}

\usepackage{natbib}
\usepackage{cleveref}

\setcitestyle{authoryear}

\begin{document}
\title{Air taxi skyport location problem with single-allocation choice-constrained elastic demand for airport access}

\author{
  \IEEEauthorblockN{Srushti Rath and  Joseph Y.J. Chow \\}
    Department of Civil and Urban Engineering\\
    New York University, NY, USA  \\
    Email: srushti.rath@nyu.edu,  joseph.chow@nyu.edu} 

\maketitle 
\begin{abstract}
 Witnessing the rapid progress and accelerated commercialization made in recent years for the introduction of air taxi services in near future across metropolitan cities, our research focuses on one of the most important consideration for such services, \emph{i.e.,} infrastructure planning (also known as skyports). We consider design of skyport locations for air taxis accessing airports, where we present the skyport location problem as a modified single-allocation p-hub median location problem integrating choice-constrained user mode choice behavior into the decision process.
 Our approach focuses on two alternative objectives \emph{i.e.}, maximizing air taxi ridership and maximizing air taxi revenue. The proposed models in the study incorporate trade-offs between trip length and trip cost based on mode choice behavior of travelers to determine optimal choices of skyports in an urban city. We examine the sensitivity of skyport locations based on two objectives, three air taxi pricing strategies, and varying transfer times at skyports. 
 A case study of New York City is conducted considering a network of 149 taxi zones and 3 airports with over 20 million for-hire-vehicles trip data to the airports to discuss insights around the choice of skyport locations in the city, and demand allocation to different skyports under various parameter settings. Results suggest that a minimum of 9 skyports located between Manhattan, Queens and Brooklyn can adequately accommodate the airport access travel needs and are sufficiently stable against transfer time increases. Findings from this study can help air taxi providers strategize infrastructure design options and investment decisions based on skyport location choices.\\
 

\end{abstract}

\section{ Introduction }
\label{sec:introduction}
Major cities around the world are currently struggling with a common problem: traffic congestion resulting from urban population growth and limited roadway capacity.
Spikes in travel time across congested routes in a city can have unpleasant consequences, \emph{e.g.}, not reaching the airport on time to catch a flight. Furthermore, in an extreme scenario, congestion can delay emergency services, \emph{i.e.}, delay in moving a critical patient to a medical center. Such concerns have pushed the transportation industry and academia to come up with new higher speed transportation modes to avoid surface congestion altogether. In this context, eVTOL (electric vertical takeoff and landing) vehicles, also known as electric 'air taxis' are emerging as a promising option to improve urban mobility. While helicopter services have been around for quite a long time in various metropolitan areas, (\emph{e.g.,} New York City, Los Angeles, São Paulo), the concept of urban air mobility (UAM) or advanced air mobility (AAM) for passenger transportation is focused on providing on-demand shared mobility using technology-efficient, less noisy, affordable, environmental friendly and potentially automated aerial vehicles (\emph{i.e.,} air taxis). While air taxi services have not yet been launched in any major city, multiple industry groups (including air manufacturers, large private companies, and smaller start-ups) are actively working on projects focused towards offering such services in the next few years (e.g., Uber, Hyundai, Toyota, EHang, Volocopter, AirBus, Boeing, Lilium Jet, Terrafugia, Joby Aviation, Kitty Hawk and others).

In 2016, the transportation network company Uber released a comprehensive white paper (\citealp{uber}) with a follow-up technical document (\citealp{uberelevate}) discussing their view on requirements of urban air taxis to make UAM feasible as an affordable solution to commuters. It is estimated that producing high volumes of safe and reliable air taxis would drive down passenger costs per trip (\citealp{Tim}; \citealp{summit}). Furthermore, various technology requirements and regulatory steps associated with on-demand aerial mobility have been discussed by NASA (\citealp{nasa, johnson}); these include the use of distributed electric propulsion, concept vehicle design, power and energy requirements, noise and emission reduction, safety, and reduction in operation and energy costs. In past years, major brands and other eVTOL start-ups made tremendous progress towards making the UAM concept a reality (\citealp{Ronan, Shane}). Recently, NASA signed Space Act Agreements with 17 companies in the aviation industry to conduct full field tests in urban environments. The UAM Grand Challenge (\citealp{UAMchallenge}) by NASA is first in a series of technology demonstrations aimed at evaluating different elements of UAM operations under various weather, traffic and contingency conditions.  A first version of UAM Concept of Operations was recently released by the Federal Aviation Administration (\citealp{FAA}) providing an initial road map on achieving high volume and safe urban air taxi operations. 
Such active interest and serious funding supporting air taxi projects is indicative of its (potential) widespread adoption in the near future.

Identifying the UAM market and understanding customers' preferences to on-demand eVTOL services is the primary consideration for planning UAM operations. As a preliminary assessment of the impact of air taxi services, several groups have conducted surveys and analyses. For example, \cite{sioux} confirmed via simulations that the reduction in travel time may strongly influence the adoption of air taxi services. Another survey conducted by Airbus (\citealp{airbus}) spanning three regions (New York City, Frankfurt and Shanghai) indicated that airport access/transfers are the best use case for UAM adoption by commuters. The white paper by Uber (\citealp{uber}) identified  infrastructure development as a key challenge in enabling efficient UAM operations. In their paper, the term 'skyport' or 'vertiport' denotes the ground infrastructures required for air taxi operations (\emph{i.e.}, boarding, alighting of passengers, eVTOL charging, take-off and landing operations). The air taxi service is multimodal in nature \emph{i.e.}, the end-to-end trip would mainly involve use of ground transportation to and from the skyports (\citealp{goyal}). \cite{Kreimeier_economic} found that the willingness to pay for such on-demand aviation services is greatly affected by the first mile and last mile ground transportation distance. Thus, location of skyports based on travelers' choices might be a crucial factor in the overall adoption of this emerging mode of transportation.

Our study draws motivation from the key findings in the above studies. We focus on the problem of planning the ground infrastructure for air taxi services for airport access/transfers in an urban city. In particular, we incorporate travelers' preferences for UAM services to determine optimal skyport locations in an urban city with an objective of providing economically sustainable solutions to air taxi service providers. Given a large city with multiple candidate skyport locations, it is non-trivial to select a small subset as skyports that attracts all customer groups with different preferences, especially when the location choices and users' decisions of using air taxi services are closely interrelated. The example described below explains these challenges in the context of a real city (New York City (NYC)).\\ 

\begin{figure}[!htb]
\begin{center}
\includegraphics[scale=.65]{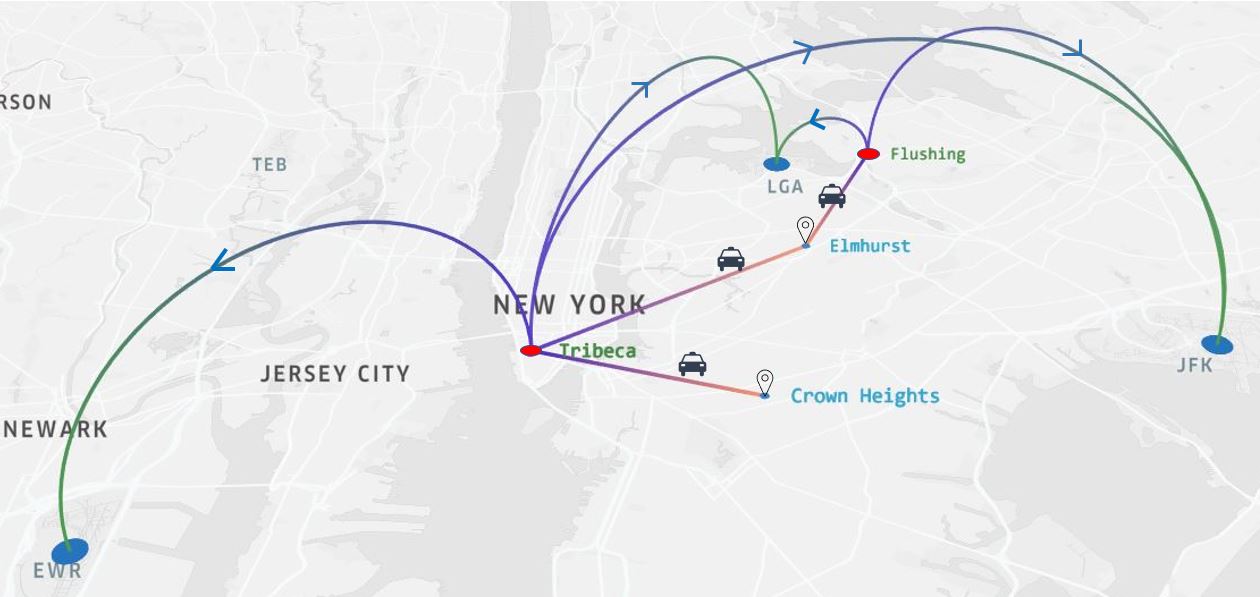}
\caption{Illustrative example showing two skyport locations in NYC (Flushing and Tribeca), and the connecting edges (ground mobility and air mobility) to nearby airports, \emph{i.e.}, JFK, EWR, and LGA for different origin locations (Elmhurst and Crown Heights). The straight lines connecting two points denote ground mobility (via ground taxi) from origin to skyport, and the arcs connecting two points depict air mobility connection (via air taxi) from skyport to airport landing zones.} \label{fig:airtaxi_demo}
\end{center}
\end{figure}

\paragraph*{Illustrative example}
Consider three major airports in NYC: John F. Kennedy International airport (JFK), Newark Liberty International airport (EWR), and LaGaurdia airport (LGA) as shown in \Cref{fig:airtaxi_demo} and two origin locations:  Elmhurst (Queens, NYC) and Crown Heights (Brooklyn, NYC), each having significant demand to the three airports. Let us assume two skyports located at Tribeca (Manhattan) and Flushing (Queens) respectively. \Cref{fig:airtaxi_demo} shows ground mobility connections from origin locations to skyports followed by air taxi trips to landing zones (located nearby respective airports). Assuming a set of decision variables (\emph{e.g.}, trip time, trip distance, trip cost etc.) influencing users' behavior of choosing air taxi service, for demand originating at Elmhurst (to LGA and JFK airports), the skyport at Flushing would be the lowest generalized cost (in travel time and fare,  inferred using real data from NYC), whereas, for commuting to EWR airport the best option is the skyport at Tribeca. On the contrary, for the airport demand originating at Crown Heights the optimal path to all the three airports is via the skyport at Tribeca. In general, while planning the location of (multiple) skyports in a city to maximize ridership or maximize revenue, one needs to consider the intricacies depicted in this example (multiple airports and allocation based on user preferences). In addition, the potential demand from each location in the city estimated to go via different skyports to multiple airports depends on the generalized costs relative to a next best alternative, \emph{i.e.} surface taxi.\\

\paragraph*{\textbf{Contribution}}
The decision of locating a skyport in a metropolitan area largely depends on the travel propensity towards air taxi services. Previous studies have highlighted the significance of cost and time in determining the potential adoption of on-demand aerial mobility (\citealp{fu_rothfeld, kreimeier,peeta}). 
The contribution of this study is a proposed skyport location problem with elastic demand response and solution methodology that can be used by air taxi service providers for determining skyport locations in urban cities considering short term and long term price scenarios. The problem is a variant of the classic hub location problem (HLP) (\citealp{kelly,camp2})
where given a set of fixed facilities (in our case airports) each serving various demand points, the location-allocation problem optimizes subsets of demand locations to be assigned to different transfer points (in our case skyports) for connection to multiple facilities (airports). We model the objective function in two ways leading to two variants of the skyport location problem \emph{i.e.}, (1) revenue maximization problem and (2) ridership maximization problem. Our method optimizes the skyport locations considering travel costs, transfer times, and user demand for air taxis using a logit model. 
Typically such a formulation would be nonconvex due to the logit model constraint, but we show that by enumerating optimal paths for each origin-skyport-airport we can 
reformulate it into a linear model that can be solved using linear programming solution approaches.
We analyze the sensitivity of skyport locations to three different pricing strategies (considering short-term and long-term price scenarios) and varying transfer times at skyports, to study how these parameters affect the choice of optimal skyport locations in a city.
To be clear, this is an operational demonstration study; profitability analysis of air taxi services is beyond the scope of this study. Although the results are demonstrated for NYC, this method is fairly general and can be applied to other cities selected for air taxi operations. The choice of an appropriate method for determining the skyport locations is important to ensure air taxi operational efficiency.

The remainder of this paper is organized as follows. Section \ref{sec:related_work} covers related work (including prior work on UAM infrastructure, and location models in the broader hub location framework) and highlights research gaps that motivated this study.
The proposed methodology is described in Section \ref{sec:methodology}.
This is followed by Section~\ref{sec:experiments} on the experiment and results and we conclude in Section~\ref{sec:conclusion}.

%%%%LITERATURE REVIEW%%%%
\section{Literature review} \label{sec:related_work}
\subsection{Prior work on eVTOL and research gaps} \label{sec:research_gap}
We first describe the recent industry research surrounding UAM and eVTOLs followed by prior work on eVTOL infrastructure planning.

Due to the rapid technological advancement, the concept of on-demand urban air mobility has captured much public attention and research interest in recent years. Electric VTOLs have been an area of active investment for various air manufacturers, service companies, and high-tech giants who have been pushing for its public acceptance. 
While the development of a safe and efficient eVTOL is necessary for implementation of UAM, various challenges and barriers to the future of UAM have been highlighted and addressed by companies, consulting services and regulatory agencies. Some of the key challenges include public acceptance, market space, airspace integration, safety, noise, emissions, air traffic management, regulations, certification, and pilot training (\citealp{thippha,McKinsey,Deloitte_insights,airbus,uber,porsche,FAA,shaheen,nasa}). Additionally, one of the crucial factors identified was ground infrastructure (skyport) selection (\cite{uber}) to enable efficient air taxi operations and link multiple modes of urban transportation for a complete end-to-end trip (\citealp{deloitte}). The multimodal structure of the air taxi service would include first mile access from origin location to the skyport, followed by boarding on the eVTOL, flight leg (take-off and landing) on a landing pad, de-boarding, and last mile transfer from the landing pad to nearest destination (\citealp{deloitte,porsche}). Due to limited space availability in urban cities, utilization of existing helipads as well as rooftops of high-rise parking garages or buildings for eVTOL operations is a topic of active interest (\citealp{uber,deloitte_new,mckinsey_infra}) targeted at reducing the air taxi infrastructure cost . 
A recent study highlighting current research and development in UAM (\citealp{straubinger2020overview}) indicates ground infrastructure as a key determinant in successful adoption of this technology.

In the academic research community, 
prior work on 
eVTOL infrastructure selection
focus on operational and space requirements for UAM infrastructure while discussing several layout options for skyports (\citealp{alexander, vascik, vascik2, vascik3, vascik4}). These studies highlight various factors that contribute toward infrastructure planning of UAM (\emph{e.g.}, population density, income levels, long commute times to work, congestion, tourism, and airport trips)
and propose suitable locations for infrastructure (\emph{e.g.},
existing helipads, rooftops of existing parking garages or high rise buildings, roadways, and open spaces in large intersections). Access to airports or other major transport hubs was ranked the highest in a study by \cite{fadhil_gis} (in terms of importance of factors in assessing ground infrastructure station selection); a GIS-based analysis was used for studying skyport placement options. A recent work by \cite{clustering}
proposed selecting skyport locations by using K-means clustering algorithm,
and demonstrated results for the Seoul metro area. The clustering of trips to determine skyport locations (cluster centroids) was limited to only three major routes within the city. In addition, there was no explicit optimization algorithm to minimize the aggregate travel time based on the trips data, and the clustering approach was essentially a heuristic. Another similar study is that of \cite{raj}. They also looked at skyport placement, starting first with a comprehensive analysis of the demand requirements for skyport siting before diving into analysis involving k-means clustering approach.  
The authors only looked at demand within 1 mile of each skyport and ignored the access/egress times in the computation of the switch to air taxi. Also, the air taxi eligible demand accounts for only the trips with air taxi travel times at least 40\% less than the ground trip duration. These approaches do not exploit the spatial structures of multimodal paths (access, flight, and egress) connecting origins to destinations through different skyports located across an urban city. Considering the importance of access and egress ground transportation in the willingness to pay for such services (\citealp{Kreimeier_economic}), it is essential to consider the trade-offs between the time savings and the service cost in the skyport planning process. 

In order to understand the user demand due to such factors (\emph{i.e.,} trip length and trip cost) and to be able to incorporate effects of such factors on user behavior in the planning process, a discrete choice based demand model can be used. Based on multiple factors (individual specific, mode specific, attitudinal, social, psychological, and latent variables), various studies have used discrete choice models to understand travel behavior and user adoption of UAM services (\citealp{balac_prospects,balac_demand,garrow2019,binder2018,roy2021user,boddupalli2019,ilahi,haddad,fu_rothfeld}).
A detailed market study by Booz-Allen and Hamilton, Inc. (\citealp{shaheen}) estimated
the potential demand of UAM in several cities in the United States, such as New York City, Los Angeles, Washington, D.c., San Francisco Bay area; we use their findings in our study. 
Using a demand model to incorporate traveler decisions and choices for UAM services, we propose a more general and principled optimization framework based on HLP for optimizing the locations of skyports in any given city. A survey by \cite{airbus} indicated that users were willing to pay more for air taxi services to avoid the negative consequences of ground transportation congestion leading to delays in airport transfers. Motivated by the survey results, we focus on airport access/transfers as a use case and optimize skyport locations for (1) maximizing air taxi ridership and (2) maximizing air taxi revenue. The advantage of modeling the skyport location problem using HLP structure with a mode choice model is that it is able to capture access costs to the skyports along with transfer costs of switching between modes, and optimize the locations based on the cost trade-offs to reflect user behavior.

\subsection{Related work on hub location problem and choice-constrained optimization} \label{sec:related_work_hlp}

The problem setup in our study has fundamental connections with HLP. Hubs serve as transfer or switching points in a many-to-many distribution network. 
The basic framework of the network HLP is to select the locations for hubs in order to serve $N$ demand points accessing a facility in a network such that they fulfill an objective (\emph{e.g.}, minimizing distance or travel time between origin-destination pairs in a network, maximizing profit from the hub facility). The potential locations for a hub facility are the trip origin locations (nodes) in the network.

The HLP can be applied to all areas where demand points need to be routed through some transfer locations or hubs (such that several demand points can be collected together at these hubs) for distribution to facilities. For example, this can be employed in an emergency aid system (\citealp{furuta}) where patients can be transferred by ambulance to heliports (hubs) to be flown to hospitals (facilities). Usually this approach is used when the travel distance (or time) from a hub to the facility is comparatively less compared to traveling directly from demand points to the facility. Therefore, we use the concepts of HLP for locating multiple skyports to connect demand points in a city to multiple airports as the facilities within the city. 

Depending on the objective, there can be three major variations to the HLP (\citealp{camp1,camp3}):
\begin{enumerate}
\item $p$-hub median (minisum),
\item $p$-hub center(minimax), and
\item covering problem,
\end{enumerate}
where $p$ is the number of hubs (typically an input parameter). In the $p$-hub median problem, the objective is to minimize the total transportation cost. This cost is defined in terms of the travel distance or the travel time from origin to destination. The $p$-hub median problem is NP-hard; even if the hub locations are fixed, the allocation part of the problem remains NP-hard (\citealp{kara1999}). The problems which include service time are typically formulated as $p$-hub center or hub covering problems. While the objective in $p$-hub center problems is to minimize the maximum distance between origin and destination (O-D) pairs, the hub covering problem focuses on maximizing the service coverage.

The first mathematical formulations of HLP were introduced by \cite{kelly}. In the HLP literature, this formulation is referred to as a single allocation $p$-hub median problem, where $p$ is the number of hubs (typically an input parameter). The first linear integer programming formulation of this quadratic model was proposed by \cite{camp2}. Various linear models for HLP were later proposed by \cite{ernst} and \cite{skorintight}. The hub facilities in the HLP are categorized as uncapacitated (where there is no capacity restriction) and capacitated (where there is a limit to the maximum flow passing through a hub) (\citealp{aykin, ebery}). Several studies consider hub location under congestion effects, where the delay of accessing a hub is dependent on the flow entering the hub. Examples include \cite{maria,de,oz}. The reader may refer to \cite{campbell2002hub,farahani,alumur} for a  comprehensive review on classification of various models and approaches to HLP. 
The classic HLP assumes the unit travel cost from a demand point to the hub is the same as that of traveling directly to the facility, whereas the unit travel cost between hubs (or from the hub to the facility (\citealp{berman2007transfer}) is reduced by a discount factor ($\beta$ \emph{i.e.,} $0 < \beta < 1 $). Some variants of HLP also allow direct connection from origin to the facility without being routed through a hub (\citealp{berman2007transfer, berman2008multiple, stochastic_tplp}). In line with these assumptions, our setup considers use of ground transportation from demand points to skyports as well as for direct connections to airports, with air taxi connections (faster speed) from skyports to airports. The discount factor and direct connections in the HLP are accounted for in our skyport location model by mode specific travel costs in the multimodal setup.

Generic HLP problems are modeled mainly with an objective to minimize total network cost to satisfy all demand. However, when the decision to allocate demand via hubs is dependent on trade-off between various decision variables (as per user behavior), it is beneficial to define the model objective accordingly. For example, from a ridership point of view, it may be more advantageous to locate skyports such that the air taxi demand (\emph{i.e.,} demand allocated to skyports) is maximized. Similarly, from a revenue perspective, the total fare collected from the demand going via skyports should be maximized. There are limited studies in the HLP literature focusing on maximization objectives. One variant of HLP with such objective is the hub maximal covering location model introduced by \cite{camp2} where hubs try to maximize the demand coverage. Various extension to this model with different notions of coverage were investigated by  \cite{kara,hamacher,tan,wagner,hwang}. \cite{alibeyg2016hub} introduced hub network design with profit and provided exact solution for such problems (\citealp{alibeyg2018exact}). The profit calculation is based on total revenue obtained from captured flows minus the total cost of establishing hubs. Variations to this model setup can be found in \cite{neamatian} and \cite{taherkhani}. 

In terms of incorporating choice behavior in network optimization models, earliest works by \cite{ andersson1989operational, andersson1998passenger, algers2001modelling} apply logit choice models to estimate buy-up and recapture factors at Scandinavian airlines system hubs. \cite{marin} incorporate user decisions as a constraint to model the rapid transit network for optimizing the location of the transit infrastructure. To address the additional complexity introduced by non-linear constraints (resulting from the use of a mode choice model), authors use piece-wise linear approximation for solving the network optimization model. The use of discrete choice models in revenue management in the context of airlines can be found in \cite{talluri2004revenue} and in mixed integer linear programs in \cite{paneque2021integrating}. 
These studies highlight the benefits and complexity of modeling user-based behavior in solving network optimization and revenue management problems under elastic passenger demand. 

The skyport location problem in our study aims to determine the optimal locations of skyports for origin-destination pairs in a network, decide which subset of origin nodes to be served by the set of skyports to multiple destinations, and make allocation decisions based on user demand model. We formulate our skyport location problem with two different objectives: (1) maximizing air taxi ridership (formulated as ridership maximization problem) and (2) maximizing air taxi revenue (formulated as revenue maximization problem). As such, the hubs are treated as uncapacitated. Rather, we incorporate a mode choice model to include elastic user demand of going through the skyports (to multiple airports) based on trade-offs between different factors (influencing user decisions). To make the problem tractable, the mode choices (and market shares) for each route are preprocessed so that they do not result in nonlinear constraints. Unlike the choice-constrained optimization problems in the literature, this is possible because we consider single allocation--all demand is assigned to a single shortest path for those taking air taxi--and the path can then be enumerated and its mode choice probability precalculated. Therefore, user choices are explicitly taken into account for non-trivial demand allocation (routing) decisions to skyports.

%%%%%METHODOLOGY%%%%%
\section{Methodology} \label{sec:methodology}
We first provide a high level overview of our approach used to formulate the skyport location problem in Section~\ref{sec:overview}; this is followed by the formal problem formulation in Section~\ref{sec:problem_formulation}.

\subsection{High level overview} \label{sec:overview}
For a given city characterized by a discrete set of locations and multiple airports associated with it, the air taxi skyport location problem is formulated as a variant of the HLP. The skyport location problem in our study is defined as an optimization model with two alternative objectives \emph{i.e.,} either maximizing air taxi ridership or maximizing air taxi revenue (not as a multi-objective problem but as two alternative optimization models). The number of skyports is a (budget) constraint in the formulation. Considering trip length and trip cost of the transportation mode as major influencing factors in our setup, we only consider a major subset of airport travelers (\emph{i.e.}, regular taxi users) and estimate the behavioral mode shift of these travelers towards air taxis using a mode choice model (\citealp{mcfadden}). The sensitivity of skyport locations to varying transfer time and trip cost is analyzed to study how these parameters affect the choice of optimal skyport locations. In this context,

\begin{itemize}
\item \textit{single allocation} refers to the constraint where demand at each origin node is satisfied via a single skyport, where users are assumed to choose between the skyport against taking direct ground transportation (there is no stochastic route choice among the skyports over which passengers are probabilistically distributed)
\item no capacity constraint is considered; our study is based on a subset of air taxi demand (\emph{i.e.,} airport demand) with explicit demand response, hence the formulation is \textit{uncapacitated} (reader may refer \cite{vascik_capacity} that studies skyport capacity),
\item no fixed \textit{infrastructure cost} are considered (number of skyports is considered as the budget constraint in our setup), 
\item conditions for \textit{reduced travel cost} from transfer points to facilities and \textit{direct allocation} of demand nodes to facilities (as in HLP) are both reflected in the mode choice model integrated in our skyport location problem that diverts a portion of the population of taxi passengers to air taxi,
\item no \textit{congestion effects} (see \citealp{de}) are assumed at the skyports, and
\item \textit{existing landing space} for helicopters in airport zones are assumed to serve as landing zones for air taxis in our setup.
\end{itemize}

To be clear regarding the conditions above, we use the HLP model to design skyport locations that account for the access, egress and transfer costs allowed in the air taxi market. The benefits of switching to a multi-leg trip are estimated using the elastic demand which forms the basis of our skyport location problem. In the remainder of this paper we use the term hub and skyport interchangeably.

\subsection{Problem formulation} \label{sec:problem_formulation}
In the following subsections, we first describe our setup with formal notation, decision variables, and then the optimization problem with associated constraints.\\

\subsubsection{Setup} \label{sec:setup}
Consider a discrete set $\mathcal{L} = \{1,2,3,\ldots,N\}$ of locations spread across a given city. We consider trips from a location $i \in \mathcal{L}$ to an airport $j \in \mathcal{J}$, where  $\mathcal{J}$ is the set of airports in the city (discrete set of size $N_{dest} \geq 1 $). In other words, we only focus on trips which have an airport in set $\mathcal{J}$ as their destination.
 
Two main modes of commute to airports are considered, \emph{i.e.,} direct ground service (ground taxi) and aerial service (air taxi), as we assume the air taxis would primarily compete with ground taxis. As such, the taxi user population is used as the market from which users may shift to air taxi. Given such alternatives, we assume an individual's choice of travel mode (to a destination) is greatly influenced by trip length and price of the trip made by the mode (\citealp{fu_rothfeld}). Since the pricing of air taxi services would be adjusted based on market conditions over time, the focus is not on the pricing decisions but how user responses to such decisions affect the choice of skyport locations. Therefore, in terms of air taxi price, the estimates by Uber (\citealp{uber_price, uber}) are assumed. As per these estimates, the air taxi service price, on a per passenger mile basis, will initially cost \$5.73 which may go down to \$1.86 before reaching to \$0.44 in the long-term (\emph{i.e.}, comparable to car ownership variable cost). Based on these values, we consider three different price scenarios for our analyses:
\begin{itemize}
    \item Short term (ST): Air taxi price is \$5.73 per passenger mile
    \item Medium term (MT): Air taxi price is \$1.86 per passenger mile
    \item Long term (LT): Air taxi price is \$0.44 per passenger mile
\end{itemize}

\begin{figure}[!htb]
\begin{center}
\includegraphics[scale=.62]{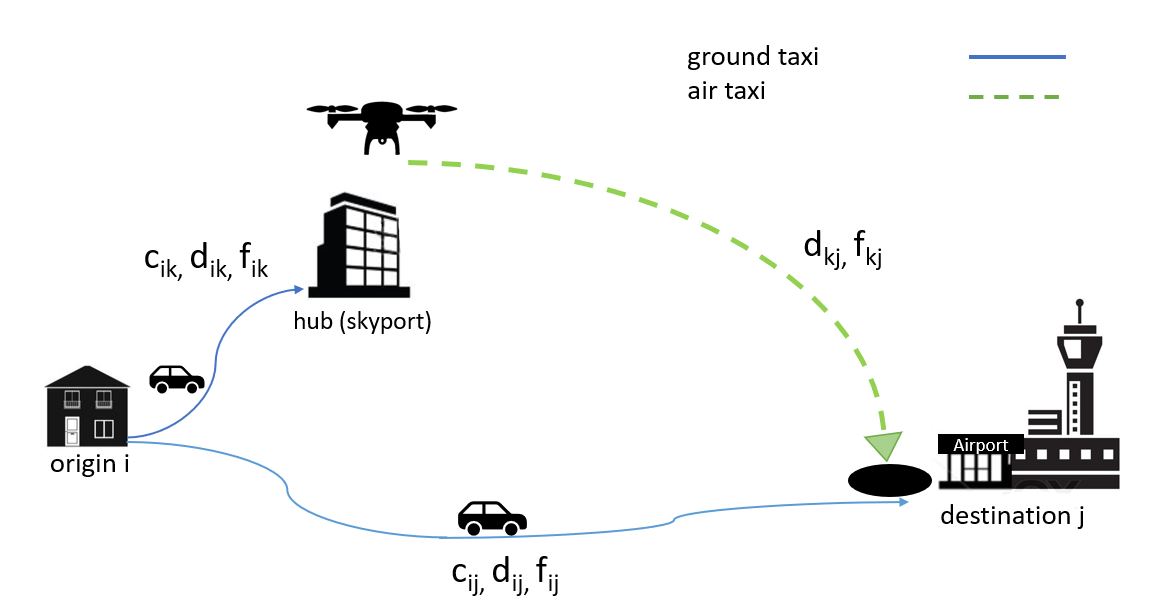}
\caption{Allocation of demand from an origin $i$ to destination airport $j$ as per commute options and decision variables. The demand can be  satisfied via a single skyport $k$ (\emph{i.e.,} $i$ allocated to $k$) using air taxi and via direct ground transportation ($i$ to $j$) using ground taxi. $c_{ab}$ indicates trip time from point a to point b, while $d_{ab}$ is the corresponding distance, and $f_{ab}$ denotes the associated trip fare}.  \label{fig:routing} 
\end{center}
\end{figure}

In our setup, the originating airport demand at each origin node is fulfilled either via a skyport or via direct ground transportation (as shown in Figure~\ref{fig:routing}); the proportion distribution depends on the mode choice decisions by users based on attributes such as trip length and trip cost to the destination airport as per commute options.
Travelers' decisions to choose air vs ground taxi from an origin to a destination (OD) are reflected by a binary logit model (as described in Section~\ref{sec: demand_model}); each OD pair has one shortest path for the air taxi model for which attributes are computed using the parameters described below.\\

\subsubsection{Parameters} \label{sec:parameters}
The set of parameters to the proposed optimization problem (as shown in Figure~\ref{fig:routing}) are defined as follows:
\begin{itemize}
\item $c_{ik}$ $\triangleq$ ground transportation (access) trip time (minutes) between origin $i$ and candidate skyport $k$ with $i,k \in \mathcal{L}$ and $c_{ik} \geq 0 $

\item $d_{ik}$ $\triangleq$ ground transportation (access) trip distance (miles) from origin $i$ to candidate skyport $k$ with
$i,k \in \mathcal{L}$ and $d_{ik} \geq 0 $

\item $c_{ij}$ $\triangleq$ direct ground transportation trip time (minutes) from origin $i$ to destination $j$
with $i \in \mathcal{L}$, $j \in \mathcal{J}$, and $c_{ij} \geq 0 $ 

\item $d_{ij}$ $\triangleq$ direct ground transportation trip distance (miles) from origin $i$ to destination $j$
with $i \in \mathcal{L}$, $j \in \mathcal{J}$, and $d_{ij} \geq 0 $ 

\item $d'_{kj}$ $\triangleq$ ground trip distance (miles) from node $k$ to destination airport $j$ with $k \in \mathcal{L} $, $j \in \mathcal{J}$, and $c_{kj} \geq  0$

\item $d_{kj}$ $\triangleq$ aerial distance (miles) from candidate skyport $k$ to airport $j$ with
$k \in \mathcal{L}$, $j \in \mathcal{J}$, and $d_{kj} \geq 0 $
\begin{itemize}
    \item[$\bullet$] $d_{kj}$ $\backsim$ f($d'_{kj}$) : explained in Section~\ref{sec:travel_costs}
\end{itemize}

\item $f_{ij}$ $\triangleq$ fare associated with direct ground transportation trip (USD\footnote{United States Dollar}) from origin $i$ to destination $j$ with $i \in \mathcal{L}$, $j \in \mathcal{J}$, and $f_{ij} > 0$ 
\begin{itemize}
    \item[$\bullet$]  $f_{ij}$ $\backsim$ f($c_{ij}$, $d_{ij}$): explained in Section~\ref{sec:travel_costs}
\end{itemize}

\item $f_{ik}$ $\triangleq$ fare associated with access ground transportation trip from origin $i$ to candidate skyport $k$ (USD) with $i,k \in \mathcal{L}$, $f_{ik} \geq 0$
\begin{itemize}
    \item[$\bullet$]  $f_{ik}$ $\backsim$ f($c_{ik}$, $d_{ik}$): explained in Section~\ref{sec:travel_costs}
\end{itemize}

\item $f_{kj}$ $\triangleq$ fare associated with air taxi commute (USD) from candidate skyport $k$ to destination $j$ with $i \in \mathcal{L}$, $j \in \mathcal{J}$, $f_{kj} > 0$
\begin{itemize}
    \item[$\bullet$] $f_{kj}$ $\backsim$ f($d_{kj}$): as mentioned in Section~\ref{sec:setup}
\end{itemize}

\item $f_{ikj}$ $\triangleq$ cost associated with end-to-end air taxi trip (USD) from origin $i$ to destination $j$ via candidate skyport $k$ with $i \in \mathcal{L}$, $k \in \mathcal{L}$, $j \in \mathcal{J}$, $f_{ikj} \geq 0$ 
\begin{itemize}
    \item[$\bullet$] $f_{ikj}$ = $f_{ik} + f_{kj}$
\end{itemize}

\item $D_{ij}$ $\triangleq$ demand originating from $i$ to destination airport $j$ with $i \in \mathcal{L}$, $j \in \mathcal{J}$, and $D_{ij} \in \mathbb{N} = \{1,2,3,4,5,\ldots\}$

\item $p$ $\triangleq$ number of skyports\\
\end{itemize}

\subsubsection{Incorporating transfer times} \label{sec:transfers}
\begin{figure}[!htb]
\begin{center}
\includegraphics[scale=.67]{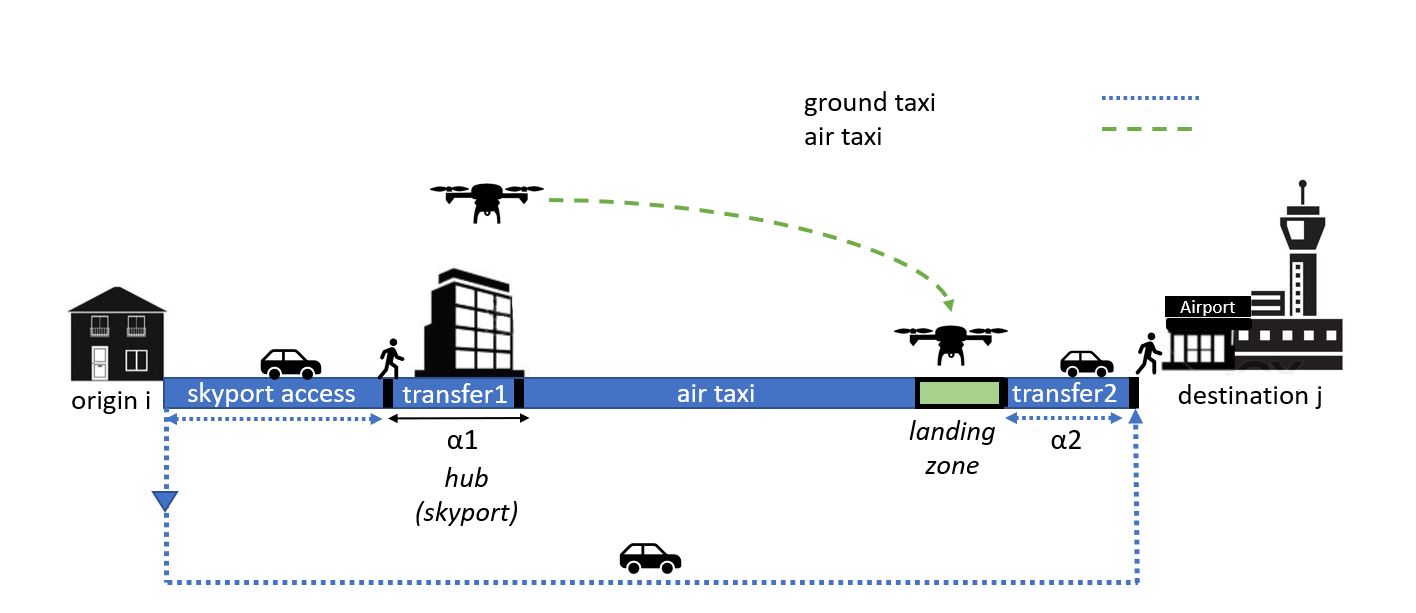}
\caption{Multimodal structure of air taxi service from origin $i$ to destination airport $j$ via a skyport $k$ involves ground access, transfer1, air taxi, and transfer2.} \label{fig:multimodal_trip} 
\end{center}
\end{figure}

UAM use is a multimodal trip (shown in Figure~\ref{fig:multimodal_trip}), which includes switch between modes. This involves some amount of time (\emph{i.e.,} additional travel cost) to transfer to and from the skyports.
 
\begin{itemize}
\item{\textit{transfer1 ($\alpha_1$)}}: this refers to the time required to switch between ground transportation to air taxi along with access time to the take off zone of the skyport; we consider an equivalent in-vehicle (ground transportation) time as $\alpha_1$,

\item{\textit{transfer2 ($\alpha_2$)}}: this refers to the last mile transfer via ground transportation from a landing zone (\emph{e.g.,} existing helipad located nearest to the destination airport) to the destination airport terminal.
\end{itemize}

In order to test the sensitivity of optimal hub locations and the demand for those hubs to transfer times, we include a transfer cost in the air taxi service. This is reflected in the mode choice model to account for the impact of transfers on users' behavior. We essentially consider a transfer cost ($t_{k}$) associated with the total transfer time ($\alpha_1$ + $\alpha_2$ $\geq 0$); this value is captured in the total air taxi cost ($f_{ikj}$) as shown in Eq. (\ref{eq:UAM_fare}).
\begin{align}
    f_{ikj} = f{ik} + t_k + f_{kj}\label{eq:UAM_fare}
\end{align}

\subsubsection{Choice constraints} \label{sec: demand_model}  We integrate choice-constrained user mode choice behavior into our optimization problem. Since we consider single allocation skyport location problem, all demand originating from a location is assigned to a single shortest path for those taking air taxi; hence we enumerate the mode choice probability (derived from a mode choice model) of each path which is incorporated into the location-allocation decision process (described below). This way user choices are explicitly taken into account to include elastic user demand of going through the skyports to multiple airports for the skyport location problem. 

Considering trip length and trip cost of the transportation mode as major influencing factors in our setup, we only consider a major subset of airport travelers (\emph{i.e.}, regular taxi users) and estimate the behavioral mode shift of these travelers towards air taxis using a binary logit model (\citealp{mcfadden}). In this model, the utility maximization rule states that the individual chooses the alternative with the highest utility. The modeling approach for a binary logit model involves defining an utility function for individual $n$, associated with alternative $a$ ($U_{n,a}$) from a choice set $\mathcal{A}$. The utility can depend on the attributes of alternatives and individual characteristics. $U_{n,a}$ has two components \emph{i.e.,} a deterministic component ($V_{n,a}$) which is the observable portion of the utility, and the error component ($\epsilon_{n,a}$) which is assumed to be Gumbel distributed.

We adopt findings available from existing studies (and surveys) on user behavior in NYC to define the utility functions of ground taxi and air taxi in our binary logit model (\cref{eq:util_groundtaxi,eq:util_airtaxi}). A detailed market survey on UAM by Booz-Allen and Hamilton, Inc. (\citealp{shaheen}) estimates a logistic regression model considering multiple individual and alternative specific variables (including UAM trip distance and UAM trip cost) to predict users preference for air taxi as a commute mode in NYC. Another study by \cite{ziyi} considers trip cost and trip time as influencing variable to estimate a mode choice model comparing taxi (or for-hire-vehicles) along with other modes for airport access in NYC. 

We do recognise that, in practice, a detailed (stated preference) survey specific to the study area need to be conducted (while considering various other factors influencing user choices) to estimate a mode choice model for the region before using it in the infrastructure planning process. However, given a mode choice model that can estimate the potential demand for air taxis, the aim of our study is to demonstrate how such a demand model can be effectively incorporated into a formal optimization process for the skyport location planning problem and analyze the sensitivity of skyport locations to varying transfer time and trip cost. 

Both the above mentioned demand studies (\citealp{shaheen, ziyi}) use survey data from NYC (which is our study area). Since we examine only the ground taxi demand and its competition with air taxi based on travel length, we can isolate the calibrated parameters for the trip disutilities with respect to the two modes and calibrate an alternative specific constant (ASC) differentiating them. We assume that the alternative specific constants for air taxi and ground taxi are the same and thus drop out, suggesting that users only differentiate between the two modes by cost, travel time, and travel distance. This is because the ASC value would simply shift the total volume but would not change with the design of our hub locations, and would not impact the design decision if we are not considering capacitated hubs. We utilize the estimations from these models to define the utility components of alternatives in our binary logit model as shown below.

\begin{itemize}
\item Ground taxi: Trip time (minutes) and trip cost (\emph{i.e.,} taxi fare in USD)
\begin{itemize}
    \item $V_{n,ground taxi} : \textbf{f}(c_{ij}, f_{ij})$ (parameters as in Section~\ref{sec:parameters})
    \item $c_{ij}, f_{ij}$ $\mapsto \beta_{TTgroundtaxi} \times c_{ij} + \beta_{COgroundtaxi} \times f_{ij}$\\
where  $\beta_{TTgroundtaxi} = \frac{1}{min}$ and $ \beta_{COgroundtaxi} = \frac{1}{USD}$\\
\end{itemize}

\item Air taxi: UAM trip distance (air miles) and UAM trip cost (\emph{i.e.,} air taxi fare in USD)
\begin{itemize}
    \item $V_{n,air taxi} : f(d_{kj}, f_{ikj})$ (parameters as in Section~\ref{sec:parameters})
    \item $d_{kj}, f_{ikj} \mapsto \beta_{TLairtaxi} \times d_{kj} + \beta_{COairtaxi} \times f_{ikj}$\\
where $\beta_{TLairtaxi} = \frac{1}{miles}$ and $\beta_{COairtaxi} = \frac{1}{USD}$
\end{itemize}
\end{itemize}
\noindent\\
Based on the above, the utility functions of ground taxi and air taxi are shown in Eqs. (\ref{eq:util_groundtaxi}) - (\ref{eq:util_airtaxi}):
\begin{align}
    U_{n,ground taxi} &=  V_{n,ground taxi} + \epsilon_{ground taxi}\nonumber\\
    \implies U_{n,ground taxi} &= \beta_{TTgroundtaxi} \times c_{ij} + \beta_{COgroundtaxi} \times f_{ij} + \epsilon_{ground taxi}\label{eq:util_general_groundtaxi}\\
    U_{n,air taxi} &=  V_{n,air taxi} + \epsilon_{air taxi}\nonumber\\
    \implies U_{n,air taxi} &=  \beta_{TLairtaxi} \times d_{kj} + \beta_{COairtaxi} \times f_{ikj} + \epsilon_{air taxi}\label{eq:util_general_airtaxi}
\end{align}
\noindent\\
Based on the studies by \cite{shaheen} and \cite{ziyi}, we assume the values for $\beta_{TTgroundtaxi}$ = 0.0313, $\beta_{COgroundtaxi}$ = $-0.0125$ (\cite{ziyi}), and $\beta_{TLairtaxi}$ = 0.018, $\beta_{COairtaxi}$ = $- 0.0213$ (\cite{shaheen}) respectively. Therefore, \cref{eq:util_general_groundtaxi,eq:util_general_airtaxi} can be expressed as:
\begin{align}
    U_{n,ground taxi} &= 0.0313 \times c_{ij} - 0.0125 \times f_{ij} + \epsilon_{ground taxi}\label{eq:util_groundtaxi}\\
   U_{n,air taxi} &=  0.018 \times d_{kj} - 0.0213 \times (f_{ikj}) + \epsilon_{air taxi}\label{eq:util_airtaxi} 
\end{align}
Applying \cref{eq:UAM_fare}, the variable $f_{ikj}$ in \cref{eq:util_airtaxi} can be expressed as the total cost of the multi modal trip (i.e, origin to skyport via ground taxi and skyport to destination via air taxi, as shown in Figure~\ref{fig:routing}) including transfer cost, as shown in \cref{eq:util_tr_airtaxi}.

\begin{align}
    U_{n,air taxi} &=  0.018 \times d_{kj} - 0.0213 \times (f_{ik}+t_k+f_{kj}) + \epsilon_{air taxi}\label{eq:util_tr_airtaxi}
\end{align}

Note that the trip length values via air taxi are smaller than the trip costs so the utility is generally negative overall, and the positive trip length (i.e., travel time and travel distance) coefficients reflect user preferences such that people farther away in distance  prefer using air taxi and longer in time prefer ground taxi.  Furthermore, both the cost variables $f_{ij}$ and $f_{ikj}$ are functions of corresponding ground trip times and trip distances (refer calculation details in Section~\ref {sec:parameters}). Therefore, in a scenario where, for a given distance the ground trip time increases due to congestion, there is a proportional increase in its trip cost which eventually affects the user choice behavior. Essentially, the air taxi demand to the skyports is based on trade-offs between trip length and trip cost based on user preferences.

Based on the mathematical structure of the binary logit model (\citealp{mcfadden}), \cref{eq:prob_airtaxi_expression,eq:prob_ground_taxi} denote the choice probabilities of air taxi and ground taxi.\\
\begin{align}
%\textcolor{black}{P_{air taxi}} &= \dfrac{1}{1 + e^{(V_{ground taxi}-V_{air taxi})}} \label{eq:prob_airtaxi_expression}\\
P_{air taxi} &= \dfrac{e^{V_{air taxi}}}{e^{V_{air taxi}} + e^{V_{ground taxi}}}\label{eq:prob_airtaxi_expression}\\
P_{ground taxi} &= 1 - P_{air taxi} \label{eq:prob_ground_taxi}
\end{align}

The choice behavior of the origin population in Eq. (\ref{eq:prob_airtaxi_expression}) can be used to estimate the aggregate air taxi demand flow originating from $i$ to destination $j$ that is routed via a skyport at location $k$. Therefore, we define population choice behavior $\theta_{ikj}$ (based on $P_{air taxi}$) as:
\begin{itemize}
    \item $\theta_{ikj}$ $\triangleq$ population choice probability of using air taxi service from origin $i$ to destination airport $j$ if routed via skyport $k$ with $i \in \mathcal{L}$, $k \in \mathcal{L}$, $j \in \mathcal{J}$, 0 $\leq$ $\theta_{ikj} \leq 1$.\\
    
 Using \cref{eq:util_groundtaxi,eq:util_tr_airtaxi} in \cref{eq:prob_airtaxi_expression}, $\theta_{ikj}$ can be expressed as \cref{eq:pop_prob_final}:

\begin{align}
   \theta_{ikj} &= \dfrac{e^{0.018\times d_{kj} - 0.0213\times (f_{ik} + t_k + f_{kj})}} {e^{0.018\times d_{kj} - 0.0213\times (f_{ik} + t_k + f_{kj})} + e^{0.0313\times c_{ij} - 0.0125\times f_{ij}}}\label{eq:pop_prob_final}
\end{align}
\end{itemize}

Eq. (\ref{eq:pop_prob_final}) can be used to estimate the probability of a regular taxi user switching to air taxi service for given attribute values. For example, based on the illustration shown in Figure~\ref{fig:airtaxi_demo}, 
for the taxi demand at Elmhurst going to LGA airport, the probability of choosing air taxi via the skyport (using (\cref{eq:pop_prob_final}) at Flushing is 0.27. This means 27 out of 100 regular taxi users at Elmhurst would choose air taxi service to go to LGA airport via Flushing skyport. On the other hand, for the same origin demand, there is 15\% chance of using air taxi service for the skyport at Tribeca.\\

Note that only one skyport is assigned to each location; under this assumption, the demand function Eq. (\ref{eq:pop_prob_final}) need not be used as a constraint within the optimization model. Instead, we exploit the structure of the problem to pre-compute all the probabilities of Eq. (\ref{eq:pop_prob_final}) to use as objective coefficients (\emph{e.g.,} with 150 (taxi) zones, 3 airports, and 10 skyports in NYC, that is 4500 values) for an integer linear programming model.\\

\subsubsection{Decision variables}
The decision variables are defined in \Cref{eq:yk,eq:xijk}.
\begin{align}
    y_k & = 
\begin{dcases}
    1           & \text{if location $k$ is a skyport}, \\
    0              & \text{otherwise}.\label{eq:yk}
\end{dcases}  \\
    x_{ikj} & = 
\begin{dcases}
   1             & \text{if demand at origin $i$ for destination $j$ is satisfied via a skyport at location $k$},\label{eq:xijk} \\
    0              & \text{otherwise}.
\end{dcases} 
\end{align}
\;\\
\subsubsection{Optimization problem} \label{sec:basic_optimization} Using \Cref{eq:pop_prob_final,eq:yk,eq:xijk}, we develop an optimization model under two different objectives. Due to the enumeration of the mode choice into an objective coefficient, we can formulate the skyport location problem as an integer linear program.\\

\begin{itemize}
    \item \textbf{RDR model} (maximize ridership): \Cref{eq:RDRobj,eq:RDRsingle_allocation_constraint,eq:RDRhub_presence_constraint,eq:RDRtotal_hubs_constraint,eq:RDRdecision_variables} optimizes the skyport locations such that the total air taxi ridership (\emph{i.e.,} total airport trips going via skyports) is maximized.
\begin{align}
\max & \sum_{i \in \mathcal{L}} \sum_{j \in \mathcal{J}} \sum_{k \in \mathcal{L}} \theta_{ikj} D_{ij} x_{ikj}  \label{eq:RDRobj} \\
s.t. \;\;\;\;
& \sum_k x_{ikj} = 1 \; \; \; \forall i \in \mathcal{L}, \;  j \in \mathcal{J}  \label{eq:RDRsingle_allocation_constraint}\\
&x_{ikj}  \leq y_k   \; \; \; \forall i,k \in \mathcal{L}, \;  j \in \mathcal{J} \label{eq:RDRhub_presence_constraint} \\
&\sum_k y_k = p \label{eq:RDRtotal_hubs_constraint} \\
&x_{ikj},y_k \in \{0,1\} \; \; \; i,k \in \mathcal{L}, \; j \in \mathcal{J}\label{eq:RDRdecision_variables}
\end{align}

\item \textbf{REV model} (maximize revenue):  \Cref{eq:REVobj,eq:REVsingle_allocation_constraint,eq:REVhub_presence_constraint,eq:REVtotal_hubs_constraint,eq:REVdecision_variables} optimizes the skyport locations such that the total air taxi revenue (\emph{i.e.,} total air taxi fare collected from the airport trips made via skyports) is maximized.
\begin{align}
\max & \sum_{i \in \mathcal{L}} \sum_{j \in \mathcal{J}} \sum_{k \in \mathcal{L}} (f_{ik}+f_{kj})\theta_{ikj} D_{ij} x_{ikj}  \label{eq:REVobj} \\
s.t. \;\;\;\;
& \sum_k x_{ikj} = 1 \; \; \; \forall i \in \mathcal{L}, \;  j \in \mathcal{J}  \label{eq:REVsingle_allocation_constraint}\\
&x_{ikj}  \leq y_k   \; \; \; \forall i,k \in \mathcal{L}, \;  j \in \mathcal{J} \label{eq:REVhub_presence_constraint} \\
&\sum_k y_k = p \label{eq:REVtotal_hubs_constraint} \\
&x_{ikj},y_k \in \{0,1\} \; \; \; i,k \in \mathcal{L}, \; j \in \mathcal{J}\label{eq:REVdecision_variables}
\end{align}
\end{itemize}
where $\theta_{ikj}$ (in \Cref{eq:RDRobj,eq:REVobj}) refers to the expression in \Cref{eq:pop_prob_final}.\\

The objective function \eqref{eq:RDRobj} maximizes the airtaxi ridership for each origin-destination pair, while the objective function \eqref{eq:REVobj} maximizes the revenue generated by air taxi ridership for each origin-destination pair. Constraints \eqref{eq:RDRsingle_allocation_constraint} and \eqref{eq:REVsingle_allocation_constraint} allow single allocation (\emph{i.e.}, each origin node that passes through a skyport is allocated to only one skyport), while constraints \eqref{eq:RDRhub_presence_constraint} and \eqref{eq:REVhub_presence_constraint}
ensure that demand for each destination $j$ at an origin node $i$ is satisfied via the node at $k$ if and only if a skyport is located at $k$. Constraints \eqref{eq:RDRtotal_hubs_constraint} and \eqref{eq:REVtotal_hubs_constraint} denote that the total number of skyports to be located is $p$. This is an indirect measure of the fixed cost; alternatively, if unit fixed costs are known exactly with respect to the other costs in the objective, they can be added to the objective as a fixed charge term. In our case, the budget constraint approach is used in combination with sensitivity analysis of the budget (see \Cref{tab:opt_results_ST,tab:opt_results_MT,tab:opt_results_LT}) to give a decision-maker the flexibility to compare costs once they know the fixed costs. For revenue calculation, we consider the air taxi operation model proposed by \cite{uber} where the end-to-end air taxi trip (\emph{i.e.,} ground transportation access to skyports and air taxi ride to airports) is provided by a single operator. Other air taxi service operators planning to provide service only from skyports to multiple destinations may have different pricing strategies. For example, the helicopter service by Uber in NYC (\citealp{ubercopter}) comprises of access and egress ground trips along with aerial ride to the airport, and charges passengers for the end-to-end journey (average cost ranges between \$200 - \$225). On the other hand, the BLADE helicopter service that transports passengers only from helipads to NYC airports costs between \$145 to \$195 (\citealp{ott}). The air taxi price values in our study is based on estimates by Uber (\citealp{uber_price}), hence the revenue in our setup includes total revenue generated from end-to-end multimodal air taxi trips.We provide additional analysis in \Cref{sec:optimization_results} which is a special case of the REV model; here the revenue calculation is based only on the air taxi rides from skyports to airports.

We implement the solver for the optimization problems in Section~\ref{sec:basic_optimization} using the Gurobi optimization tool (\cite{gurobi}).
For integer programming problems, Gurobi uses multiple solution methods to obtain an exact solution, \emph{e.g.},
parallel branch-and-cut algorithms, non-traditional tree-of-trees search algorithms,
cutting plane methods, and symmetry detection.
\subsection{Illustrative example}
\begin{figure}[ht]
\begin{center}
\includegraphics[scale=.65]{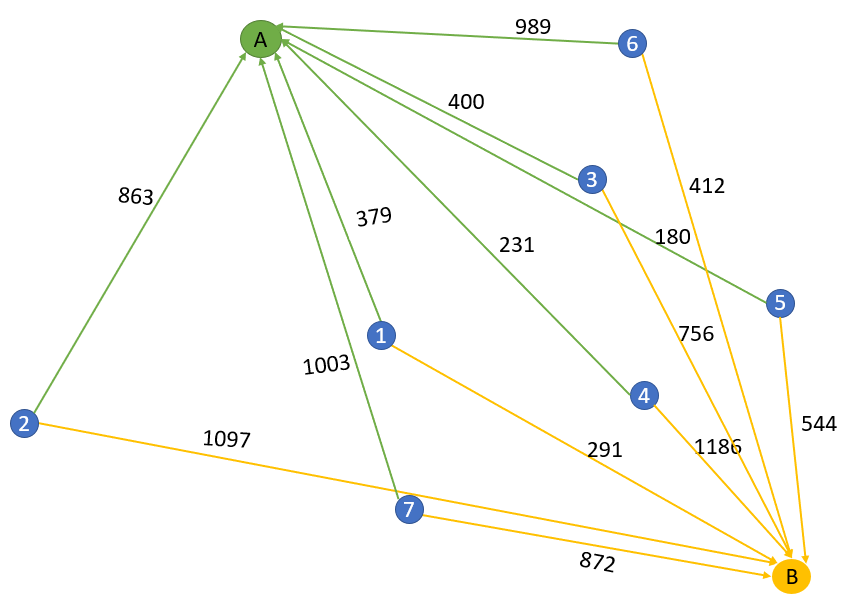}
\caption{Illustrative example showing 7 origin locations and 2 destinations; each link displays aggregate demand going from origin to destination via direct connection} \label{fig:demand_example}
\end{center}
\end{figure}

To illustrate the properties of RDR and REV models, the following example is used.  
Consider 7 origin locations (in set $\mathcal{O}$: \{1,2,3,4,5,6,7\}) and 2 destination facilities (in set $\mathcal{D}$: \{A,B\}) distributed in a Euclidean space. Each origin has direct connection to each destination ((\emph{e.g.,} ground taxi connections); the aggregate demand from origin to destination via direct connections  are shown in \Cref{fig:demand_example}. The objective is to select 2 skyports (among 7 candidate skyport locations in $\mathcal{S}$: \{1,2,3,4,5,6,7\}) to connect each origin to destination facilities via these skyports. \Cref{tab:prob_example} shows probability of users (in $\mathcal{O}$) choosing to go via a skyport (in $\mathcal{S}$) to reach a destination (in $\mathcal{D}$), while \Cref{tab:fare_example} shows the total fare associated with such trips. The value corresponding to row $i$ (\emph{e.g.}, $O1$) and column $k-j$ (\emph{e.g.}, $S1-B$) in \Cref{tab:prob_example} represents the choice probability of users in origin $i$ to go via a skyport at $k$ to reach destination $j$ (\emph{e.g.}, $O1-S1-B$ = 0.66); \Cref{tab:fare_example} follows a similar notation. The values displayed in \Cref{tab:prob_example,tab:fare_example} are only for illustration purpose in this example. The calculation of choice probabilities and travel costs used in the study are described in \Cref{sec:experiments}. 

%%%%% Probabilities
\begin{table}[ht]
\caption {Air taxi choice probabilities of users \emph{i.e.,} user probabilities of choosing to go via skyports in $\mathcal{S}$ from origin locations $\mathcal{O}$ to destinations $\mathcal{D}$: for illustrative example} \label{tab:prob_example}
\centering
\begin{tabular}{l | c | c | c | c | c | c | c |c | c | c | c |c | c | c }
\hline
	&	S1-A	&	S2-A	&	S3-A	&	S4-A	&	S5-A	&	S6-A	&	S7-A	&	S1-B	&	S2-B	&	S3-B	&	S4-B	&	S5-B	&	S6-B	&	S7-B\\
\hline
O1	&	0.66	&	0.59	&	0.51	&	0.39	&	0.52	&	0.51	&	0.32	&	0.66	&	0.47	&	0.44	&	0.58	&	0.3	&	0.56	&	0.33\\
O2	&	0.58	&	0.66	&	0.42	&	0.35	&	0.41	&	0.21	&	0.5	&	0.63	&	0.62	&	0.37	&	0.4	&	0.44	&	0.38	&	0.46\\
O3	&	0.65	&	0.41	&	0.67	&	0.47	&	0.45	&	0.51	&	0.53	&	0.48	&	0.55	&	0.62	&	0.32	&	0.34	&	0.52	&	0.51\\
O4	&	0.6	&	0.53	&	0.53	&	0.68	&	0.31	&	0.4	&	0.39	&	0.39	&	0.37	&	0.54	&	0.68	&	0.54	&	0.44	&	0.35\\
O5	&	0.42	&	0.31	&	0.43	&	0.52	&	0.64	&	0.56	&	0.46	&	0.47	&	0.61	&	0.45	&	0.58	&	0.61	&	0.32	&	0.4\\
O6	&	0.5	&	0.57	&	0.37	&	0.45	&	0.55	&	0.65	&	0.31	&	0.39	&	0.57	&	0.45	&	0.65	&	0.58	&	0.69	&	0.43\\
O7	&	0.49	&	0.36	&	0.41	&	0.4	&	0.35	&	0.39	&	0.59	&	0.55	&	0.54	&	0.4	&	0.46	&	0.34	&	0.42	&	0.52\\
\hline
\end{tabular}
\end{table}

\begin{table}[ht]
\caption {Total fare (USD) associated with air taxi trip from an origin in $\mathcal{O}$ to a destination $\mathcal{D}$ via skyport in $\mathcal{S}$ including access cost from origin to skyport and air taxi cost from skyport to destination: for illustrative example}  \label{tab:fare_example}
\centering
\begin{tabular}{l | c | c | c | c | c | c | c |c | c | c | c |c | c | c }
\hline
	&	S1-A	&	S2-A	&	S3-A	&	S4-A	&	S5-A	&	S6-A	&	S7-A	&	S1-B	&	S2-B	&	S3-B	&	S4-B	&	S5-B	&	S6-B	&	S7-B\\
\hline
O1	&	180	&	178	&	181	&	185	&	189	&	126	&	104	&	111	&	122	&	156	&	162	&	141	&	155	&	191\\
O2	&	115	&	118	&	112	&	184	&	114	&	100	&	166	&	119	&	115	&	186	&	138	&	142	&	100	&	174\\
O3	&	124	&	173	&	149	&	195	&	193	&	134	&	117	&	155	&	129	&	164	&	103	&	124	&	131	&	171\\
O4	&	149	&	166	&	195	&	101	&	145	&	191	&	135	&	133	&	179	&	173	&	111	&	196	&	175	&	105\\
O5	&	114	&	145	&	190	&	184	&	101	&	160	&	183	&	105	&	168	&	121	&	160	&	155	&	198	&	187\\
O6	&	103	&	193	&	200	&	180	&	194	&	162	&	125	&	192	&	172	&	117	&	157	&	178	&	182	&	189\\
O7	&	123	&	140	&	132	&	120	&	130	&	170	&	151	&	118	&	104	&	169	&	157	&	173	&	124	&	142\\
\hline
\end{tabular}
\end{table}

\begin{figure}[!htb]
\begin{center}
\includegraphics[scale=.62]{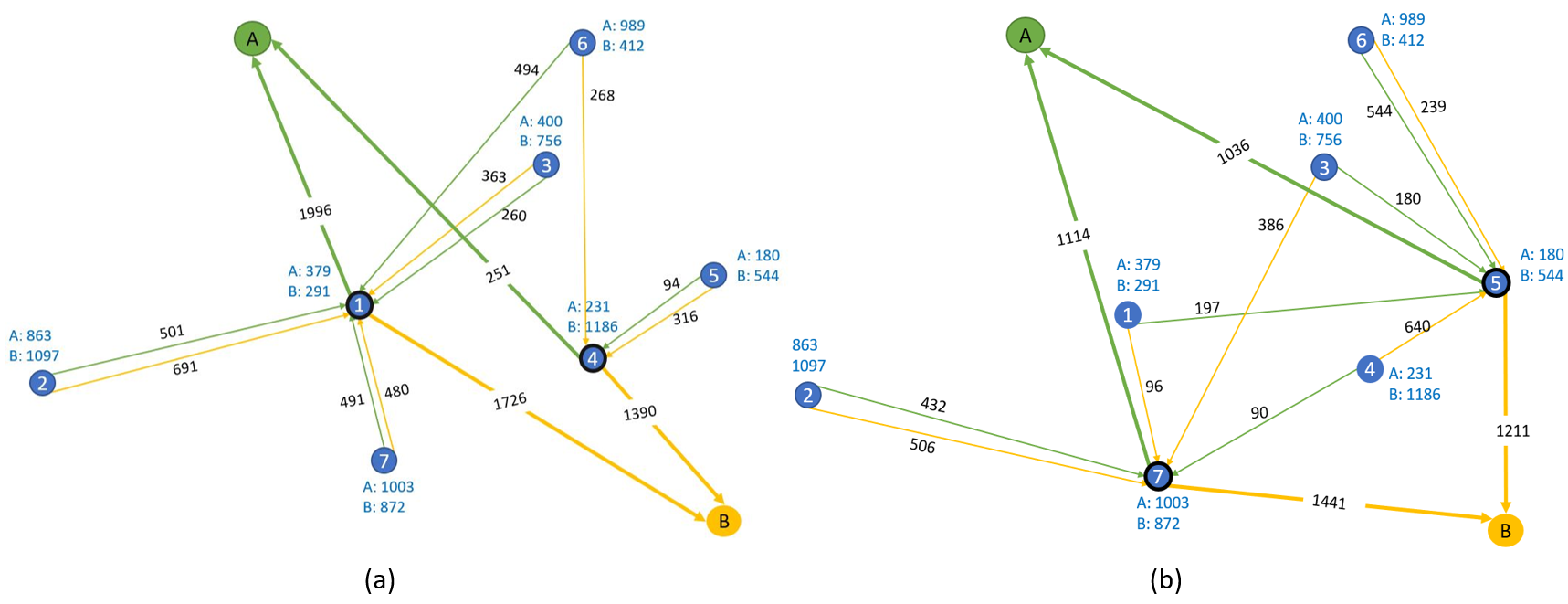}
\caption{Comparison of location-allocation solutions of RDR model in (a) and REV model (b). The figure shows allocation of air taxi demand from each origin to the selected skyport locations (highlighted in black outlines); links are color coded as per destination choices. Values on links connecting origin $i$ to skyport $k$ denote air taxi riders originating from $i$ going via skyport $k$ to their respective destinations, while those connecting skyport $k$ to destination $j$ is the aggregate air taxi demand allocated to $k$ to be transport to $j$.} \label{fig:example_modelsetup}
\end{center}
\end{figure}

Using the values of demand, choice probabilities, and fare in this example, we define the parameters used in the optimization models (\Cref{sec:basic_optimization}) to determine optimal skyport locations. In this case, $p$ = 2; \Cref{tab:prob_example} and \Cref{tab:fare_example} provide $\theta{ikj}$ and $f_{ik}+f_{kj}$ values respectively (where $i$ denotes origin, $k$ is skyport and $j$ denotes destination). The values of parameter $D_{ij}$ are as per the numbers shown in \Cref{fig:demand_example} for origin $i$ to destination $j$. Therefore, for a candidate skyport in set $\mathcal{S}$, for example at location $3$ ($S3$), the demand from origin at $2$ ($O2$) to destination $A$ via skyport $S3$ is calculated by multiplying the user probability $O2-S3-A$ from \Cref{tab:prob_example} by the aggregate demand from location $2$ to $A$ in \Cref{fig:demand_example}. Therefore, 363 riders (0.42 $\times$ 863) from $O2$ would choose to go via skyport $S3$ to destination $A$; the revenue generated from these air taxi riders is obtained by multiplying the ridership with corresponding air taxi fare for $O2-S3-A$ in \Cref{tab:fare_example} which gives $\$40,656$ (363 $\times$ 112). Also for $O2$, this means 500 out of 863 users directly go to $A$ without using skyports. Similarly the air taxi demand from different origin to each candidate skyport (going to destinations $A$ and $B$) are computed to be used in the location-allocation decision process for obtaining an optimized set of skyport locations based on the objective in Section~\ref{sec:basic_optimization}.  
Substituting the parameter values considered in this illustrative example in the RDR model (\cref{eq:RDRobj,eq:RDRsingle_allocation_constraint,eq:RDRhub_presence_constraint,eq:RDRtotal_hubs_constraint,eq:RDRdecision_variables}), the optimal choice of skyports are at location $1$ and $4$, whereas the REV model (\cref{eq:REVobj,eq:REVsingle_allocation_constraint,eq:REVhub_presence_constraint,eq:REVtotal_hubs_constraint,eq:REVdecision_variables}) result in optimal skyport locations at $5$ and $7$. The skyport locations and allocation of origin demand to the selected skyports for each destination (obtained using Gurobi) are shown in \Cref{fig:example_modelsetup}. Therefore, the total air taxi ridership (demand) at selected skyports (shown in \Cref{fig:example_modelsetup}) obtained by RDR model is 5363 compared to 4802 by REV model. The total estimated revenue from potential air taxi riders in this example are \$677831 and \$742621 as obtained from RDR and REV models respectively. 

\newpage
\section{Data set and experimental results} \label{sec:experiments}
We first describe the NYC data sets and tools used for our experiments, and then go over the results of our approach.

\subsection{Data set and tools}
\subsubsection{Data set}\label{sec:dataset}
The study area includes five boroughs in NYC (The Bronx, Manhattan, Queens, Brooklyn and Staten Island) that are divided into $263$ taxi zones (\cite{taxi_zone}); each taxi zone has a unique zone ID.
The centroids of the taxi zones are the trip origin nodes (locations) while the trip destinations are the three major airports 
in NYC, \emph{i.e.}, EWR (Newark), JFK and LGA (LaGuardia).
The NYC taxi and limousine commission FHV (for-hire-vehicles) trip record data
from 2019 (July to December) are used (\cite{TLC}) on our study.
This is a publicly available data set, and the motivation for using FHV trips data
for airport transfer is derived from the study by \cite{ziyi}. Theri study indicated that $65\%$ of trips to JFK are via FHV and taxis.

Each trip record includes the origin and destination locations of the trip in terms of taxi zone IDs.
For example, taxi zone IDs for Newark, JFK and LaGuardia airports are: $1$, $132$ and $138$ respectively. The data set contains over $20$ million trip records
and each trip record includes the start time stamp, origin taxi zone ID, end time stamp,
and destination taxi zone ID for the trip. All the trips with trip time (\emph{i.e.,} end time stamp - start time stamp) greater than 120 minutes were disregarded from the dataset along with other outliers. Furthermore, to avoid distortion in average or aggregate values, major holidays such ad Independence day, Halloween, Thanksgiving, Christmas were filtered out before performing any calculations.

\begin{figure}[!htb]
\begin{center}
\includegraphics[scale=.60]{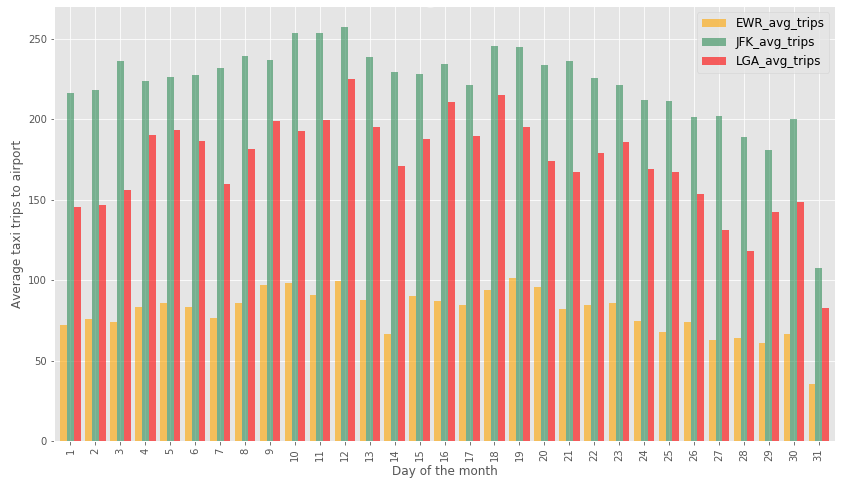}
\caption{Daily (average) taxi demand to three major airports in NYC (originating from all taxi zones)} \label{fig:airport_demand}
\end{center}
\end{figure}

Assuming the skyport operations are likely to schedule from 7 am to 6 pm (\citealp{goyal}), we only consider the trips during this period for demand calculation. 
The total airport demand and travel costs were calculated using a script written in Python programming language (version 3.7.4).
For each origin-destination ID pair in the FHV data set, the total trips were added to obtain the total demand from the 
origin ID to the destination ID.
The trips with destinations as airports (\emph{i.e.}, with destination taxi zone IDs $1$, $132$ and $138$) were selected and used to compute the demand $D_{ij}$ from origin $i$ to destination airport $j$. We found the average monthly trips originating from each of the taxi zones to the three airports; we refer to this as $D_{ij}$. Figure~\ref{fig:airport_demand} shows the profile of daily (average) taxi trips to the three airports in NYC.\\

\subsubsection{Data pruning} \label{sec:data_pruning}
\begin{figure}[!htb]
\begin{center}
\includegraphics[scale=.53]{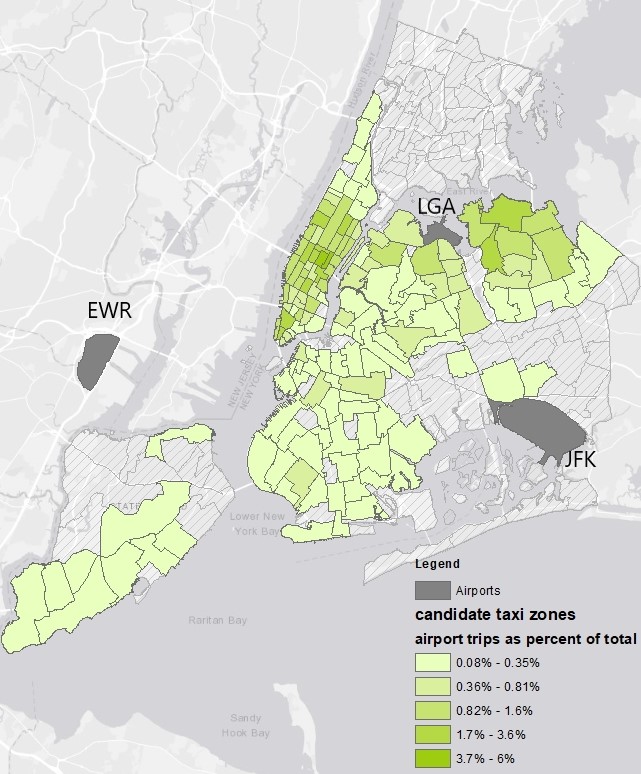}
\caption{Choropleth map showing the candidate taxi zones with taxi zone-wise demand for the three airports in NYC with zone shading.} \label{fig:choropleth}
\end{center}
\end{figure}

To further prune the data set, we filter out taxi zones which do not have significant demand to airports. The filtering process was implemented using the demand data $D_{ij}$ (as obtained in Section~\ref{sec:dataset}).  For our experiments, we excluded the taxi zones with the lowest fraction of trips (\emph{i.e.,} zones with less than 10 airport trips in a month) and as a result focused on the remaining $149$ taxi zones (\emph{i.e.}, the top $149$ zones contributing to the airport demand). Figure~\ref{fig:choropleth} shows a choropleth map of the candidate taxi zones generated using ArcMap version 10.5.1 (\cite{arcgis}). Hence, as per the notation defined in Section~\ref{sec:setup}, $|\mathcal{L}| = 149$ and $|\mathcal{J}| = 3$ (\emph{i.e.}, three airports).\\

\subsubsection{Calculation of travel costs}\label{sec:travel_costs}
The input to the skyport location problem is composed of total trips to each airport from each taxi zone ($D_{ij}$) and the associated trip travel costs across taxi zones (including airports). The travel costs include trip time, trip distance and trip cost (\emph{i.e.,} trip fare). For extreme case scenario, we consider weekday rush hours in NYC (\emph{i.e.,} 4 - 7 pm)\footnote{https://www1.nyc.gov/site/tlc/passengers/taxi-fare.page\#} to reflect on congested conditions for calculating the parameters defined in Section~\ref{sec:parameters}.\\

The average peak hour ground travel time ($c_{ik}, c_{kj}$) and average ground trip distance ($d_{ik}, d'_{kj}, d_{ij}$) across taxi zones and airports were found using Google Maps Distance Matrix API and Python programming language; values were obtained during 5pm-6pm (peak hours) on a weekday. 
The aerial distance ($d_{kj}$) was calculated based on the relation between ground miles and aerial miles (\cite{goyal}):
\begin{align}
Ground miles &= 1.42 \times Air miles\nonumber\\
\implies d'_{kj} &= 1.42 \times d_{kj}\nonumber\\
\implies d_{kj} &= \dfrac{d'_{kj}}{1.42}\label{eq:air_miles}
\end{align}

For trip fare calculations, we consider the ground transportation fare structure (used by taxi services in NYC) associated with trip time and trip distance values; the cost components include a Booking fee ($Base fee$), a per mile cost ($R_{mile}$), and a per minute cost ($R_{minute}$). Therefore, the  ground taxi (access) fare from origin to candidate skyport ($f_{ik}$) and direct ground taxi fare from origin to airport ($f_{ij}$) can be written as:
\begin{align}
    f_{ik} = Base fee + R_{mile} \times d_{ik} + R_{minute} \times c_{ik}\label{eq:access_fare}\\
    f_{ij} = Base fee + R_{mile} \times d_{ij} + R_{minute} \times c_{ij} \label{eq:direct_fare}
\end{align}
\noindent
Based on average market rates in NYC (\cite{uber_fare_NYC}), we consider the following values for $f_{ik}$ and $f_{ij}$:
\begin{itemize}
    \item $Base fee$ = \$3
    \item $R_{mile}$ = \$1.5
    \item $R_{minute}$ = \$0.3
    \item Minimum ground taxi fare between taxi zones: \$7 
    \item Minimum fare for ground taxi trip with pick up or drop off location in Manhattan: \$8 and an additional congestion charge of \$2.75 (\cite{congestion_charge})
\end{itemize}
Using the above values in \Cref{eq:access_fare,eq:direct_fare}, the access taxi fares to skyports and direct ground taxi fares to airports were computed for different values of trip times and trip distances (for each origin zone).\\

The flight leg fare of an air taxi trip \emph{i.e.,} $f_{kj}$ was determined by multiplying the aerial distance $d_{kj}$ (Eq \ref{eq:air_miles}) by the price per air mile. It is possible that for passengers sharing the same destination, an air taxi serves more than one passenger at a time, but for worst case analysis, we consider average passenger occupancy as 1. Therefore, assuming one air mile is equivalent to one passenger mile, let the price per air mile be represented as $R_{airmile}$ such that:
\begin{align}
    f_{kj} = R_{airmile} \times d_{kj}\label{eq:flight_leg_fare}
\end{align}
Based on ST, MT and LT scenarios (as explained in  Section~\ref{sec:setup}), the values of $R_{airmile}$ include:
\begin{itemize}
    \item $R_{airmile}$ = \$5.73 for short term
    \item $R_{airmile}$ = \$1.86 for medium term
    \item $R_{airmile}$  = \$0.44 for long term
\end{itemize}

Using \Cref{eq:access_fare,eq:flight_leg_fare}, the total air taxi fare $f_{ikj}$ in Eq (\ref{eq:UAM_fare}) consisting of access fare, transfer cost, and flight leg fare can be written as:
\begin{align}
    f_{ikj} &= f_{ik} + t_k + f_{kj}\nonumber\\
    \implies f_{ikj} &= Base fee + R_{mile} \times d_{ik} + R_{minute} \times c_{ik} + t_k + R_{airmile} \times d_{kj}\nonumber\\
    \implies f_{ikj} &= Base fee + R_{mile} \times d_{ik} + R_{minute} \times c_{ik} + R_{minute} \times  (\alpha_1 + \alpha_2) + R_{airmile} \times d_{kj}\nonumber\\
    \implies f_{ikj} &= Base fee + R_{mile} \times d_{ik} + R_{minute} \times c_{ik} + R_{minute} \times  (tt) + R_{airmile} \times d_{kj}\nonumber\\
     \implies f_{ikj} &= Base fee + R_{mile} \times d_{ik} + R_{minute} \times (c_{ik} + tt) + R_{airmile} \times d_{kj}\label{eq:final_UAM_fare}
\end{align}
where $tt$ = $\alpha_1 + \alpha_2$ is the equivalent in-vehicle time for transfers to and from skyports.\\
For the total transfer time ($tt$), we base our assumption of $\alpha_1$ on previous findings. For example, the transfer time for rail transit was found to be valued at approximately 8 minutes of in-vehicle time (\cite{wardman}). For the transfer time $\alpha_2$ \emph{i.e.,} the last mile transfer between a landing zone (located nearest to a destination airport) and the destination airport terminal, we consider
existing helicopter services in NYC (\citealp{bladeheli,ubercopter}) that use helipads near airports for landing and transferring passengers from helipads to the airport terminals via ground taxi or shuttle (\citealp{heliservice}). For example, the average ground taxi time from heliport at JFK airport to the JFK terminal is between 5 to 8 minutes\footnote{https://www.google.com/maps/dir/Heliport+at+JFK+Airport,+Queens,+NY/JFK+Airport,+Queens,+NY}. Therefore, we assign $tt$ as 15 minutes (\emph{i.e.,} $alpha_1$ = 8 minutes and $\alpha_2$ = 7 minutes). Using the values of $tt$ and $R_{airmile}$ in 
\Cref{eq:final_UAM_fare} we computed air taxi fares from each origin $i$ to each candidate skyport location $k$ ($i$, $k$ $\in$ $\mathcal{L}$) to each destination airport $j$ ($j$ $\in$ $\mathcal{J}$) for different scenarios \emph{i.e.,} ST, MT, LT (to be used for demand estimation and revenue calculation in the optimization models explained in following Section~\ref{sec:optimization_results}).\\

\subsubsection{Tools}
To solve the proposed optimization problems (as described in Section~\ref{sec:basic_optimization}), Gurobi optimization tool (\cite{gurobi})  and Python programming language (version 3.7.4) were used. Based on $149$ taxi zones and $3$ airports, the optimization problem for NYC consisted of $66752$ binary variables and zero continuous variables. Our experiments were carried out on a computer with Intel i$7$ processor with $2$ cores, $4$ logical processors and $16$ GB RAM with an average computation time below $10$ seconds (for solving an instance of the optimization problem in Gurobi for the above data set). The low computation time is likely because the model itself is uncapacitated, which we explain earlier in Section~\ref{sec:overview} with the formulation choice.

\subsection{Optimization results} \label{sec:optimization_results}
\begin{figure}[!htb]
\begin{center}
\includegraphics[scale=.70]{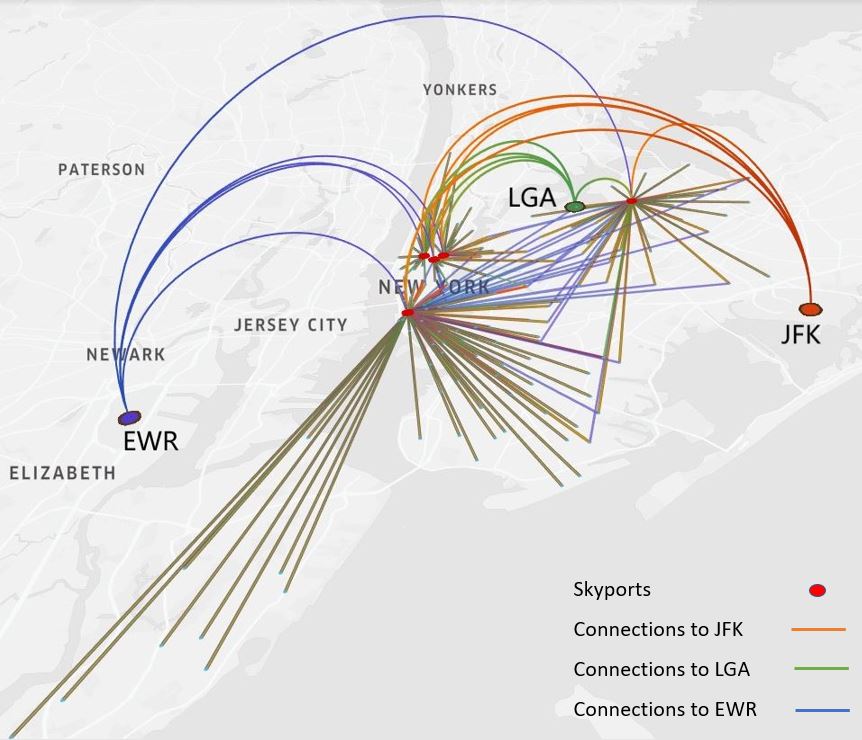}
\caption{Demand allocation showing connecting edges between origin  taxi zones to three airports in NYC via $p=5$ skyports (output of RDR model for long term scenario). The edges connecting trip origin locations to skyports are color coded as per destinations.} \label{fig:map} 
\end{center}
\end{figure}

Using the data obtained for the $149$ taxi zones to the $3$ airports (as explained above), we pre-computed all possible values of population air taxi choice probabilities ($\theta_{ikj}$) in \Cref{eq:pop_prob_final} for each origin zone to each candidate skyports (going to multiple airports). Using the obtained values, we solved the proposed optimization models \emph{i.e.,} RDR and REV  (described in Section~\ref{sec:basic_optimization}). The objective value of RDR model provides average monthly air taxi ridership, while the REV model objective value gives the average monthly air taxi revenue. We used these models to find optimal skyport locations for three price scenarios (refer Sections~\ref{sec:setup} and \ref{sec:travel_costs}). For each case, we obtained results (using our Gurobi implementation) for different values of $p \in \{1,2,3,4,5,6,7,8,9,10\}$ (where $p$ is the number of skyports). Figure~\ref{fig:map} demonstrates the demand allocation from different origin to multiple airports via optimal skyports in NYC (obtained for the case of p = $5$ in RDR model for LT scenario)\footnote{Visualization was done using the kepler.gl tool.}. The edges connecting origin to skyport (for multiple airports) as shown in the figure are the outcome of the optimization models (\emph{i.e.,} $x_{ikj}$).

%\newpage
Using the optimal skyport locations, it is possible to estimate the ridership for REV model and revenue for RDR model respectively. The air taxi ridership corresponds to the total trips to airports originating from different origin taxi zones
that are routed via optimal skyports from REV model ($k_{REV}^{*}$). This is calculated by multiplying population choice probability $\theta_{ikj}$ (from the demand model) with airport trips $D_{ij}$ where $k$ = $k_{REV}^{*}$. Similarly, the air taxi revenue for the optimal skyports from RDR model ($k_{REV}^{*}$) is computed by multiplying air taxi ridership with the revenue generated from the multimodal trips ($f_{ik} + f_{kj}$) for $k$ = $k_{REV}^{*}$. \Cref{tab:opt_results_ST,tab:opt_results_MT,tab:opt_results_LT} summarize optimization results obtained from RDR and REV models for short term, medium term and long term price scenarios; details of optimal skyport locations are reported in \Cref{sec:appendix}. The ridership and revenue values in the results are average monthly estimates. For each choice of skyport budget $p$, the data in \Cref{tab:opt_results_ST,tab:opt_results_MT,tab:opt_results_LT} include the following:
\begin{itemize}
\item Air taxi market share: this denotes the proportion of regular taxi users that are estimated to switch to air taxi service for airport access
\item Flight leg revenue share: the total revenue generated from end-to-end air taxi trips in our study includes revenue from ground transportation access to skyports plus revenue from air taxi rides (this constitutes the flight leg of the multi-modal trip). Hence, the flight leg revenue share refers to the percentage revenue generated only from the flight leg ($f_{kj}$) with respect to the total revenue ($f_{ik}+f_{kj}$)
\item Increment in total revenue: this represents percentage increment in total revenue with increasing skyport budget $p$ (increment is measured with respect to baseline $p$ = 1)
\end{itemize}

%%%%% SHORT TERM
\begin{table}[!htbp]
\centering
\caption {Optimization results for short term scenario}  \label{tab:opt_results_ST}
\begin{tabular}{c | c | c | c |  c |  c | c}
%\begin{tabular}{*7c}
\hline
Budget &  \multicolumn{3}{c}{RDR model} & \multicolumn{3}{c}{REV model}\\
\hline
number of    & Airtaxi        &  Flight leg       &     Increment          & Airtaxi       &  Flight leg       &     Increment\\ 
skyports     &  market share  &    revenue share  &     in total revenue   & market share  &    revenue share  &     in total revenue \\
(p)           &    (\%)       &      (\%)         &      (\%)               &    (\%)      &      (\%)         &      (\%)   \\
\hline
1	&	13.21	&	60.14	&	 --	   &	11.87	&	72.31	&	 --	    \\
2	&	14.95	&	38.15	&	-5.12	&	12.82	&	63.71	&	3.70\\
3	&	16.22	&	39.14	&	-1.79	&	13.48	&	69.77	&	4.97\\
4	&	16.73	&	44.98	&	-2.49	&	13.57	&	72.22	&	5.80\\
5	&	16.92	&	49.31	&	-1.13	&	13.64	&	76.42	&	6.58\\
6	&	17.07	&	48.28	&	-1.64	&	13.92	&	78.25	&	7.29\\
7	&	17.21	&	52.01	&	-0.38	&	14.16	&	78.66	&	7.97\\
8	&	17.32	&	52.39	&	0.05	&	14.29	&	79.38	&	8.53\\
9	&	17.42	&	54.92	&	0.92	&	14.30	&	79.73	&	8.98\\
10	&	17.52	&	56.32	&	1.34	&	14.34	&	80.19	&	9.38\\
\hline
\end{tabular}
\end{table}

%%%%% MEDIUM TERM
\begin{table}[!htbp]
\centering
\caption {Optimization results for medium term scenario}  \label{tab:opt_results_MT}
\begin{tabular}{c | c | c | c |  c |  c | c}
\hline
Budget &  \multicolumn{3}{c}{RDR model} & \multicolumn{3}{c}{REV model}\\
\hline
number of    & Airtaxi        &  Flight leg       &     Increment          & Airtaxi       &  Flight leg       &     Increment\\ 
skyports     &  market share  &    revenue share  &     in total revenue   & market share  &    revenue share  &     in total revenue \\
(p)           &    (\%)       &      (\%)         &      (\%)               &    (\%)      &      (\%)         &      (\%)   \\
\hline
1	&	22.22	&	47.10	&	--		&	15.39	&	43.91	&	 --	       \\
2	&	24.19	&	50.75	&	-7.65	&	15.29	&	56.88	&	7.53\\
3	&	25.04	&	54.10	&	-9.46	&	16.27	&	58.34	&	8.15\\
4	&	25.62	&	56.45	&	-10.78	&	16.35	&	58.58	&	8.36\\
5	&	26.02	&	58.23	&	-11.50	&	16.11	&	58.46	&	8.55\\
6	&	26.38	&	59.45	&	-12.24	&	16.21	&	58.78	&	8.66\\
7	&	26.68	&	60.91	&	-13.02	&	16.33	&	58.78	&	8.76\\
8	&	26.98	&	61.73	&	-14.00	&	16.31	&	58.78	&	8.80\\
9	&	27.25	&	62.97	&	-14.17	&	16.40	&	58.96	&	8.83\\
10	&	27.50	&	64.11	&	-15.16	&	16.41	&	59.00	&	8.86\\

\hline
\end{tabular}
\end{table}

\begin{table}[!htbp]
\centering
\caption {Optimization results for long term scenario}  \label{tab:opt_results_LT}
\begin{tabular}{c | c | c | c |  c |  c | c}
%\begin{tabular}{*7c}
%\toprule
\hline
Budget &  \multicolumn{3}{c}{RDR model} & \multicolumn{3}{c}{REV model}\\
%\midrule
\hline
number of    & Airtaxi        &  Flight leg       &     Increment          & Airtaxi       &  Flight leg       &     Increment\\ 
skyports     &  market share  &    revenue share  &     in total revenue   & market share  &    revenue share  &     in total revenue \\
(p)           &    (\%)       &      (\%)         &      (\%)               &    (\%)      &      (\%)         &      (\%)   \\
\hline
1	&	27.55	&	17.57	& --			&	12.77	&	11.90	&	    --	    \\
2	&	29.37	&	20.13	&	-12.65	&	14.18	&	14.40	&	4.94\\
3	&	30.67	&	24.34	&	-16.18	&	15.99	&	14.99	&	7.11\\
4	&	31.33	&	26.09	&	-19.77	&	18.29	&	16.96	&	7.69\\
5	&	31.80	&	27.46	&	-22.32	&	18.54	&	17.05	&	8.16\\
6	&	32.22	&	29.03	&	-24.28	&	18.56	&	16.91	&	8.31\\
7	&	32.59	&	30.15	&	-25.70	&	18.35	&	16.62	&	8.41\\
8	&	32.90	&	31.11	&	-27.56	&	18.29	&	16.61	&	8.47\\
9	&	33.21	&	32.36	&	-29.11	&	18.35	&	16.60	&	8.50\\
10	&	33.50	&	33.36	&	-30.88	&	18.37	&	16.64	&	8.52\\
\hline
%\bottomrule
\end{tabular}
\end{table}

\begin{figure}[!htb]
\begin{center}
\includegraphics[scale=.55]{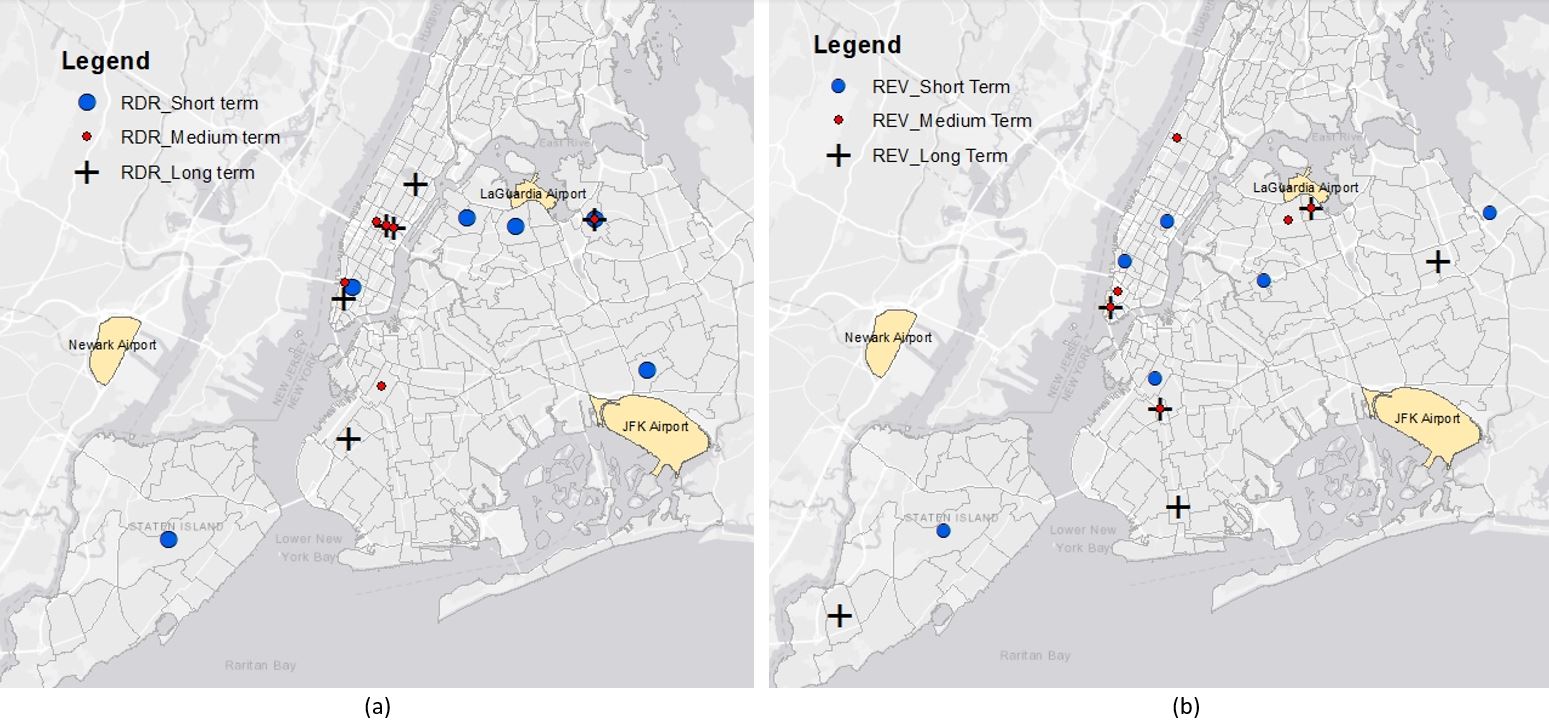}
\caption{Optimal skyport locations in NYC (for p = $6$) for short term, medium, and long term scenarios. (a) RDR model output (b) REV model output} \label{fig:REV_RDR_visualize} 
\end{center}
\end{figure}

\newpage
The output from our experiments clearly depict the sensitivity of skyport locations to different objectives as well as varying price scenarios. To get a sense of the placement of skyports, a visualization in Figure~\ref{fig:REV_RDR_visualize} shows optimal skyport locations (for $p$ = 5) in NYC obtained from RDR and REV models for ST, MT, and LT scenarios. While it can be seen that the location choice is sensitive to the objective values considered in the models, there are common zones that are optimal across different scenarios which can guide investment decisions for setting up skyports. For example, for $p$ = 6 (as shown in \ref{fig:REV_RDR_visualize}), the common skyport zones in NYC include midtown Manhattan and Flushing (Queens) from RDR model, and Park slope (Brooklyn) and Elmhurst (Queens) from REV model. Lower Manhattan area seems optimal for both RDR and REV model. 

Moreover, going from short term to long term, it can be seen that with each additional skyport in the budget, the air taxi market share in both RDR model and REV model monotonically increases, although the proportion of market share obtained from RDR model skyports is higher compared to REV model skyports. This is because the objective of RDR model is to maximize the air taxi ridership, thereby resulting in higher values of air taxi market share. Similarly, based on the objective, the total estimated revenue generated from REV model skyports are relatively higher. However, the REV model results in an increment in total revenue from additional skyports, while an opposite trend is observed in the RDR model output (especially for MT and LT scenarios, and at lower values of $p$ for the ST scenario). In this context, REV model shows relatively better performance; there is increment in both revenue and air taxi market share with increasing skyport budget $p$ (which is desired). Furthermore, for the ST scenario, the REV model output shows major portion of the total revenue being generated from the flight leg of air taxi trips; the value decreases for MT and LT scenarios due to the lower air taxi prices considered in these scenarios. However, given the higher air taxi market share estimated during the long term, various revenue management strategies can be used to ensure higher revenue from these services (\citealp{chiang,bitran}).
As mentioned earlier, we assume a single air taxi operator to provide ground taxi access to skyports with air taxi rides from skyports to airports. For cases where an air taxi operator plans to provide only air taxi rides from skyports to multiple destinations, the revenue calculation used in the REV model objective can be modified accordingly. We discuss this special case of REV model below:\\

\paragraph*{Special case (REV model)}\label{sec:special_case} 
The air taxi revenue considered in the REV model objective (Eq. \ref{eq:REVobj}) includes $f_{ik}+f_{kj}$ \emph{i.e.,} total estimated revenue from end-to-end multi-modal air taxi trips. Let this objective be denoted as $obj_{org}$. For the special case, we assume the operator is interested in maximizing only the flight leg revenue (based on the type of air taxi service it plans to provide). Hence, the revenue ($f_{ik}+f_{kj}$) in Eq. \ref{eq:REVobj} can be replaced by $f_{kj}$ (keeping constraints in \Cref{eq:REVsingle_allocation_constraint,eq:REVhub_presence_constraint,eq:REVtotal_hubs_constraint,eq:REVdecision_variables} the same). We refer this modified objective as $obj_{mod}$. As discussed earlier, based on the service type, the air taxi operators may use different pricing strategies, however, for comparison purpose (for ST scenario) we assume the same pricing strategy considered for the multi-modal air taxi setup in our study. The choice of optimal skyport locations obtained from solving the REV model with $obj_{mod}$ is different from those obtained using $obj_{org}$. However, higher number of common skyport locations were found in both objectives for higher values of $p$. The REV model with $obj_{mod}$ resulted in 7\% lower air taxi ridership for (averaged across different choices of $p$ $\in$ $\{1,2,3,\ldots ,10\}$). The flight leg revenue generated using $obj_{mod}$ is comparatively higher (10\% on an average) for $p \leq 4$, however, a decreasing trend (with revenue lift of 3-4\%) was noticed for higher values of $p$. Assuming $obj_{mod}$ in the multi-modal air taxi setup (as considered in our study), the total revenue generated from $obj_{mod}$ (by adding ground transportation access revenue with flight leg revenue from $obj_{mod}$) was found to be 2\% less than the total revenue obtained using $obj_{org}$.

\subsection{Demand distribution at skyports} \label{sec:skyport_demand_distribution}
\begin{figure}[!htb]
\begin{center}
\includegraphics[scale=.60]{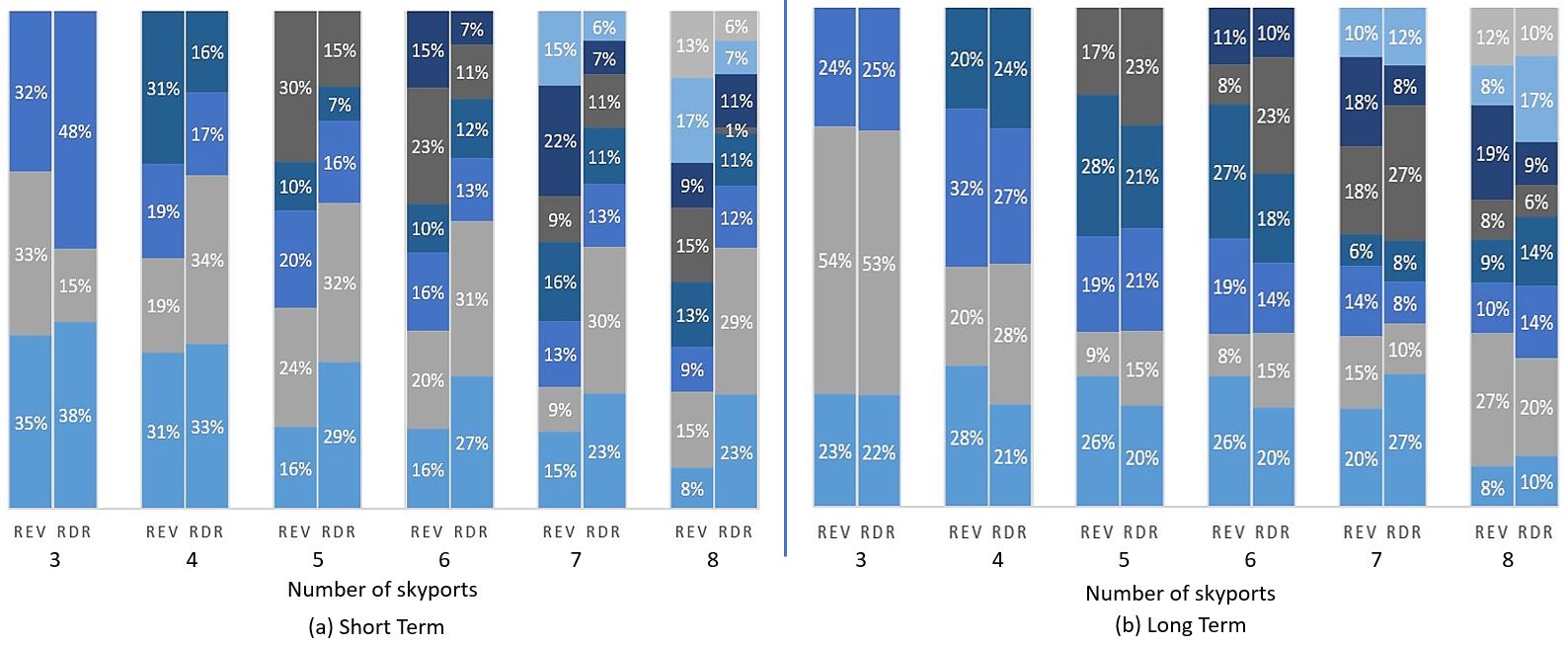}
\caption{Airport demand distribution at optimal skyport locations in NYC. Each set of bars represents percentage of demand allocated to each skyport location (given budget of $p$ skyports) obtained from REV and RDR models.} \label{fig:skyport_demand}
\end{center}
\end{figure}

As an outcome of the optimized skyport locations, we estimate the (incoming) demand at each skyport; this demand corresponds to the trips to airports
that are routed via a skyport from origin zones (allocated to that skyport). In this subsection, we study the
characteristics of such demand vis-a-vis the number of skyports.
In particular, given a budget ($p$) and the optimal skyport locations ($k$) obtained from RDR and REV models for ST and LT respectively, we compute the total number of (allocated) incoming trips to each skyport. The demand distribution at different skyports is computed using the connections between origin zones and skyports (for each airport) obtained as an outcome of the optimization models (see Section~\ref{sec:optimization_results}). These values are then used to calculate the percentage (proportion) demand allocated to each optimal skyport location (with respect to total estimated air taxi demand flow via set of optimal skyports). For the case of $p$ = $\{3,4,5,6,7,8\}$, Figure~\ref{fig:skyport_demand} shows the percentage of airport demand allocated to each skyport. As can be seen in the figure, during the short term, with increase in the skyport budget ($p$), some of the skyports in the RDR model have very less incoming demand relative other locations. For example, for $p$ = 8, the proportion of incoming trips at one of the skyports is only 1\%. For the same scenario (\emph{i.e.,} ST), the REV model skyport locations show relatively fair distribution of air taxi demand. In the long term, however, the proportion of air taxi demand across different skyports is noticed to be fairly distributed in case of both RDR and REV model. 

Overall, comparing the outcomes from RDR model and REV model, the performance of REV model is observed to be relatively better in terms of fair demand distribution at skyports and higher revenue generation (see Section~\ref{sec:optimization_results}). The distribution of air taxi demand to different skyport locations based on the choice of the optimization model (and its output) can guide service providers to plan multiple skyport design options. For instance, based on demand at optimal locations, companies can opt for smaller skyports (\emph{e.g.}, accommodating one to three air taxis) at locations with limited demand or they can choose to build skyports with higher capacity and additional facilities at locations with significant  incoming demand.

\subsection{Sensitivity analysis (time savings)} \label{sec:sensitivity_analysis}
We investigate the effects of varying total transfer time $tt$ to analyze minimum skyport requirements that can accommodate such variations. As the time spent in transfers is a critical part in the end-to-end trip of the air taxi service, we base the decision criterion of our analysis on time savings (\emph{i.e.,} time saved by travelers commuting via air taxi compared to ground taxi while going to the same destination).
As discussed earlier in \Cref{sec:transfers}, we consider a transfer cost $t_k$ associated with transfer time $tt$; this has a direct impact on user choice behavior and may affect the location allocation of skyports. The transfer time mainly depends on the infrastructure design as well as on the integration of multi modal operations by the service provider; depending on these factors the time spent in transfers can vary. Therefore, it is important to understand the minimum skyport requirements to accommodate these variations. Our analysis is limited to near term only, mainly because such variations are more likely to occur during the short term or initial period while investigating proper integration of different modes and operations design of such new services. Moreover, determining the minimum requirements in the long term would require considering various other factors beside time savings such as competitive services, user experiences, and market evolution of UAM in a city.

\begin{figure}[!htb]
\begin{center}
\includegraphics[scale=.70]{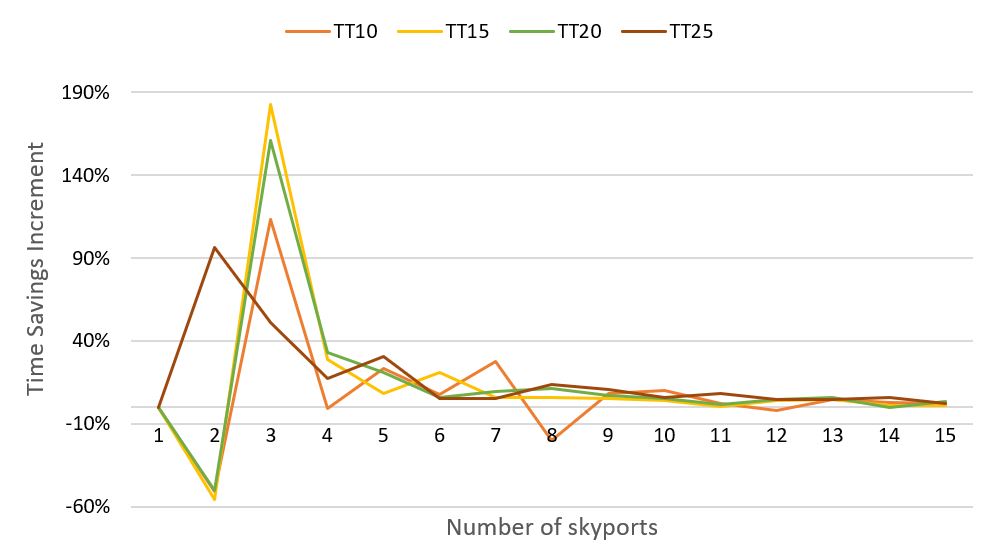}
\caption{Comparison of percentage increment in time savings ($ts_{inc}$) with respect to each additional skyport in budget $p$ (for different choices of transfer time in $TT$).} \label{fig:time_savings}
\end{center}
\end{figure}

For our analysis, we consider the REV model (because of its superior performance) and assume four possible values of $tt$, \emph{i.e.} $10$, $15$, $20$ and $25$ minutes. A study by \cite{garcia} suggests the first transfer in a multimodal trip to be equivalent to 15.2 minutes of the trip in-vehicle time. Using this finding and the values assumed in \Cref{sec:travel_costs}, we consider different choices of $\alpha_1$ (in range 8 - 15 minutes) and $\alpha_2$ (in range 5 - 8 minutes) to define the transfer time set $TT$ = \{10, 15, 20, 25\} for our sensitivity analysis. For each choice of transfer time in $TT$ and skyport budget $p$ (where $p$ $\in$ $\{1,2,3....14,15\}$), we compute the travel cost values (refer \Cref{sec:travel_costs}) and solve the REV model to get $p$ optimal skyport locations. This is used to calculate the total time savings of trips allocated to the selected skyports (for each $p$). Using these values we compute the percentage increment in time savings ($ts_{inc}$) with each additional skyport in budget $p$. \Cref{fig:time_savings} illustrates the variation of $ts_{inc}$ with respect to $p$ for different choices of $tt$ $\in$ $TT$. As shown in the figure, the variation in $ts_{inc}$ (for a given value of $p$) across different choices of  $tt$ is significantly higher for lower values of $p$. Such \textit{sensitivity} to $tt$ variations decreases as $p$ increases (with minimum variations observed at $p$ $\geq$ $9$). Therefore, based on the analysis above, the infrastructure planning for air taxi services (in the near term) requires locating at least 9 skyports across NYC. As per the REV model results (obtained for varying $tt$), the optimal choice includes one skyport each in Union Square, Financial district, Diamond district, Flushing, Murray Hill,  Park Slope, Theater District, and 2 skyports near midtown Manhattan.

\subsection{Discussion and Insights} \label{sec:insights}
Our experiments based on the formulated air taxi ridership maximization and revenue maximization optimization models (\emph{i.e.,} RDR and REV models) lead to several valuable insights that may be of interest to air taxi service operators that are planning to start their service in a metropolitan city.

\begin{itemize}
\item For a given skyport budget ($p$), the optimal set of skyports that seem attractive in the short term may not be useful for generating enough revenue in the long run. For a value of $p$, the common zones that are optimal across different price scenarios can guide infrastructure cost allocation decisions. For example, in NYC (refer \Cref{fig:REV_RDR_visualize}), the RDR model results for $p$ = 6 (for short term, medium term and long term scenarios) show common skyport zones near lower Manhattan, midtown Manhattan, and Flushing (Queens). It would make sense to allocate more infrastructure budget to design high capacity skyports in these zones.  

\item It is likely that the service operators, in the near term, would want to gain more riders (to improve familiarity with the technology and increase user adoption). In this context, the initial set of skyports can be planned based on the RDR model. However, the RDR model (for short term scenario) performs well only upto a limited skyport budget (\emph{i.e.,} for $p$ $\leq$ $p_t$). This is because for higher values of $p$ (\emph{i.e.,} $p$ $>$ $p_t$), the proportion of trips allocated to some skyports is very less and this may not be able to compensate for high infrastructure costs on such locations. For example, in case of NYC scenario for p $>$ 7 (\emph{i.e.,} $p_t$ = 7), the percentage of incoming trips at some skyports are very low; for $p$= 8, one skyport has only 1\% demand allocated to it ((see \Cref{sec:skyport_demand_distribution})). 

\item From revenue aspect, the RDR model is not greatly beneficial in the long run compared to REV model. This is because in the long run, the RDR model output does not result in higher revenue with increase in skyport budget ($p$) although it results in increased ridership (refer \Cref{tab:opt_results_ST,tab:opt_results_MT,tab:opt_results_LT}).  

\item In order for large scale air taxi operations to be sustainable in the long run, the operators would be more interested in planning for maximizing the revenue. In this context, REV model approach can be used determine the skyport choices. The fair demand distribution across different skyports obtained from REV model (see \Cref{sec:skyport_demand_distribution}) would ensure promising returns on infrastructure investments at such locations. 

\item Variation in time savings based on different choices of transfer time could serve as a possible decision criterion (besides budget constraint) for deciding minimum skyport requirements in a city. For example, as per the sensitivity analysis conducted based on time savings (see \Cref{sec:sensitivity_analysis}), it was found that atleast $9$ skyports are required in NYC to accommodate variation in transfer times.
\end{itemize}
The above insights with respect to the incorporation of elastic demand for short term and long term analyses of optimal skyport location choices in a city can facilitate the air taxi infrastructure investment decision and planning process, however, a more in-depth analysis with consideration of various other user factors in the demand model and operational factors in the optimization models is suggested as future work.

\section{Conclusion} \label{sec:conclusion}
We formulated the skyport location optimization problem for access to special destinations like airports as a variant of HLP while incorporating elastic demand toward air taxi services. Our linearized formulation can be readily solved using commercial software like Gurobi for scenarios like that of NYC; this can be easily extended to include healthcare facilities, sports venues, and major transportation hubs in addition to airports.
We formulated an optimization model with two alternative objectives in our study \emph{i.e.,} RDR model with an objective to maximize air taxi ridership, and REV model with an objective to maximize air taxi revenue. Depending on the objective, the model allocates demand from different origin zones across optimal skyports (for each destination). The demand to the skyports is based on trade-offs between trip length and trip cost based on user preferences. These preferences are incorporated in the skyport location problem using a binary mode choice logit model. We consider different price scenarios (estimated by Uber for air taxi services) in our analysis such as short term, medium term and long term.

The case study of NYC was presented considering airport access/transfers as a use case for air taxis. Using a dataset from NYC TLC with over $20$ million FHV trips to major airports associated with NYC (\emph{i.e.}, JFK, EWR, and LGA), we obtained the optimal locations for skyports for air taxis using RDR and REV models for each price scenario. Choosing the right objective and approach to locate skyports has a great impact on potential air taxi ridership as well as on the returns gained from such services. 

The choice of skyport location and allocation of demand at each skyports based on user behavior was used to study the demand distribution at each skyports; it was found that the REV model results in fair distribution of demand  across skyports compared to RDR model. Also, from revenue aspect, the REV model shows relatively better performance. 

Another important insight is around the choice of minimum number of skyports. In this context, we considered variation in travel time savings (increment) across different choices of transfer time as a decision criterion. The number of skyports in a city should be able to handle variations in transfer time; this is reflected via increase in travel time savings with respect to skyport budget ($p$). The results of sensitivity analysis for NYC (based on optimal skyport locations using REV model) show that at least $9$ skyports would ensure good performance in the near term.

%%%%%%----------------------------
Although our study is based on a (major) subset of demand, and selected (significant) decision variables reflecting user preferences, the method used in the analysis of optimal skyport locations considering short term and long term scenarios can help air taxi providers pursue various skyport design options based on infrastructure location choices. In terms of future directions, the research can be further refined and elaborated along the following lines:
\begin{itemize}
\item the optimization problem can be augmented by considering different modes of access to skyports (\emph{e.g.}, bike, walk, e-scooters, public transit) and adding other eligible (mostly long commute) trips for potential demand estimation,
\item updating demand model with other competitive modes (see \citealp{fu_rothfeld,competitiveness}) and with additional influencing factors (\emph{e.g.}, individual specific decision variables capturing changes in perceptions in post-pandemic scenario),
\item considering aerial (flight) restrictions, regulations, and noise related factors pertaining to operations in cities as per the city structure,
\item considering access to medical facilities and daily long commutes as additional factors driving UAM adoption,
\item incorporating queue management using pricing schemes,
\item modeling stochasticity in travel time and demand,
\item using simulation based modeling approaches (see \citealp{rothfeld2018agent}) to study overall performance ,
\item considering capacity effects at skyports (see \citealp{vascik_capacity}); one way would be to assign travelers to skyports with nonbinding capacities, which would also consider potential for transfers between hubs,
\item developing robust optimization methods with additional heuristics, and
\item queue sensitive air taxi rebalancing (motivated by similar work for ground transportation by \cite{relocation}).
\end{itemize}

Finally, although our results are focused on NYC, the methods are easily applicable to other cities as well.

%\section*{Acknowledgement}
%The authors wish to thank the C2SMART University Transportation Center at NYU for its support for their research project.
\newpage
\bibliography{bibliography2} 

\appendix \label{sec:appendix}
\Cref{tab:opt_locations_ST,tab:opt_locations_MT,tab:opt_locations_LT} report optimal skyport locations obtained from our experiments (for different pricing scenarios). 
%%%%% SHORT TERM
\begin{table}[!htbp]
\centering
\caption {Optimal skyport locations in NYC for short term scenario}  \label{tab:opt_locations_ST}
\begin{tabular}{c | c  | c }
\hline
Budget &  \multicolumn{1}{c}{RDR model} & \multicolumn{1}{c}{REV model}\\
\hline
number of   &  skyport locations  &       skyport locations    \\ 
  skyports (p)    &  (taxi zone IDs)   &    (taxi zone IDs)      \\
\hline
1	&	[157]	&	[170]\\
2	&	[10, 260]	&	[198, 234]\\
3	&	[10, 114, 129]		&	[114, 160, 161]\\
4	&	[7, 10, 92, 211]	&	[114, 161, 162, 198]	\\
5	&	[7, 10, 92, 161, 211]	&	[92, 161, 162, 181, 249]\\
6	&	[7, 10, 92, 114, 129, 161]		&	[92, 161, 162, 181, 230, 231]\\
7	&	[7, 10, 92, 114, 129, 161, 162]		&	[92, 161, 162, 170, 181, 230, 231]	\\
8	&	[7, 10, 92, 114, 118, 129, 161, 162]	&	[48, 92, 161, 162, 170, 181, 230, 231]\\
9	&	[7, 10, 92, 114, 118, 129, 161, 162, 230]	&	[48, 92, 161, 162, 170, 181, 230, 234, 261] \\
10	&	[7, 10, 25, 92, 114, 118, 129, 161, 162, 230]	&	[48, 92, 161, 162, 170, 181, 230, 234, 239, 261 \\
\hline
\end{tabular}
\end{table}

%%%%% MEDIUM TERM

\begin{table}[!htbp]
\centering
\caption {Optimal skyport locations in NYC for medium term scenario}  \label{tab:opt_locations_MT}
\begin{tabular}{c | c  | c }
\hline
Budget &  \multicolumn{1}{c}{RDR model} & \multicolumn{1}{c}{REV model}\\
\hline
number of   &  skyport locations  &       skyport locations    \\ 
  skyports (p)    &  (taxi zone IDs)   &    (taxi zone IDs)      \\
\hline
1	&	[170]	&		[257]\\
2	&	[92, 161]	&	[70, 87]	\\
3	&	[92, 125, 161]	&		[70, 231, 236]	\\
4	&	[92, 125, 161, 162]	&		[70, 231, 236, 257]	\\
5	&	[92, 125, 161, 162, 230]	&	[7, 24, 70, 231, 257]\\
6	&	[92, 125, 161, 162, 181, 230]	&	[7, 24, 70, 231, 257, 261]	\\
7	&	[92, 125, 161, 162, 181, 230, 236]		&	[7, 24, 70, 231, 243, 257, 261]\\
8	&	[92, 125, 161, 162, 170, 181, 230, 236]	&		[7, 24, 70, 129, 231, 243, 257, 261]\\
9	&	[48, 92, 161, 162, 170, 181, 230, 231, 236]	&	[7, 24, 70, 129, 143, 231, 243, 257, 261]\\
10	&	[7, 48, 92, 161, 162, 170, 181, 230, 231, 236]	&	[7, 24, 70, 118, 129, 143, 231, 243, 257, 261] \\
\hline
\end{tabular}
\end{table}

\begin{table}[!htbp]
\centering
\caption {Optimal skyport locations in NYC for long term scenario}  \label{tab:opt_locations_LT}
\begin{tabular}{c | c  | c }
\hline
Budget &  \multicolumn{1}{c}{RDR model} & \multicolumn{1}{c}{REV model}\\
\hline
number of   &  skyport locations  &       skyport locations    \\ 
  skyports (p)    &  (taxi zone IDs)   &    (taxi zone IDs)      \\
\hline
1	&	[170]	&	[204]\\
2	&	[92, 161]&	[135, 204]\\
3	&	[92, 161, 231]		&	[135, 204, 261]	\\
4	&	[92, 161, 162, 231]		&	[70, 135, 204, 261]	\\
5	&	[92, 161, 162, 230, 231]	&	[70, 123, 135, 204, 231]\\
6	&	[92, 161, 162, 230, 231, 236]	&	[70, 123, 135, 204, 257, 261]	\\
7	&	[92, 161, 162, 227, 230, 231, 236]	&	[70, 98, 123, 135, 204, 257, 261]	\\
8	&	[48, 92, 161, 162, 227, 230, 231, 236]	&	[44, 70, 98, 123, 135, 204, 257, 261]\\
9	&	[48, 92, 161, 162, 227, 230, 231, 236, 252] 	&	[44, 70, 98, 118, 123, 135, 204, 257, 261] 	\\
10	&	[48, 92, 161, 162, 170, 227, 230, 231, 236, 252]	&	[44, 70, 98, 118, 123, 135, 204, 231, 257, 261] \\
\hline
\end{tabular}
\end{table}

\end{document}